\documentclass[letterpaper]{article} 
\usepackage{aaai25}  
\usepackage{times}  
\usepackage{helvet}  
\usepackage{courier}  
\usepackage[hyphens]{url}  
\usepackage{graphicx} 
\usepackage{natbib}  
\usepackage{caption} 
\usepackage{amsmath}
\usepackage{algorithmic}
\usepackage{algorithm}
\usepackage{array}
\usepackage[caption=false,font=normalsize,labelfont=sf,textfont=sf]{subfig}
\usepackage{textcomp}
\usepackage{url}
\usepackage{verbatim}
\usepackage{graphicx}
\usepackage{cite}

\usepackage{amssymb}
\usepackage{booktabs}

\usepackage{enumitem}
\usepackage{times}
\usepackage{epsfig}
\usepackage{amsfonts}
\usepackage{bm}
\usepackage{multirow}
\usepackage{makecell}
\usepackage{bbding}
\usepackage[table]{xcolor}
\usepackage{tikz}

\newcommand{\ignore}[1]{}
\definecolor{cGreen}{RGB}{100,180,100}
\definecolor{cRed}{RGB}{220,50,0}

\title{SUTrack: Towards Simple and Unified Single Object Tracking}
\author{
    Xin Chen\textsuperscript{\rm 1},
    Ben Kang\textsuperscript{\rm 1},
    Wanting Geng\textsuperscript{\rm 1},
    Jiawen Zhu\textsuperscript{\rm 1},
    Yi Liu\textsuperscript{\rm 2},
    Dong Wang\textsuperscript{\rm 1}\thanks{Corresponding author.},
    Huchuan Lu\textsuperscript{\rm 1}
}
\affiliations{
    \textsuperscript{\rm 1}	Dalian University of Technology,\textsuperscript{\rm 2} Baidu Inc.\\
    chenxin3131@mail.dlut.edu.cn,
    kangben@mail.dlut.edu.cn,
    gengwanting@mail.dlut.edu.cn,
    jiawen@mail.dlut.edu.cn,
    liuyi22@baidu.com,
    wdice@dlut.edu.cn,
    lhchan@dlut.edu.cn
}

\begin{document}

\maketitle

\begin{abstract}
In this paper, we propose a simple yet unified single object tracking (SOT) framework, dubbed SUTrack. It consolidates five SOT tasks (RGB-based, RGB-Depth, RGB-Thermal, RGB-Event, RGB-Language Tracking) into a unified model trained in a single session.
Due to the distinct nature of the data, current methods typically design individual architectures and train separate models for each task. This fragmentation results in redundant training processes, repetitive technological innovations, and limited cross-modal knowledge sharing. In contrast, SUTrack demonstrates that a single model with a unified input representation can effectively handle various common SOT tasks, eliminating the need for task-specific designs and separate training sessions. Additionally, we introduce a task-recognition auxiliary training strategy and a soft token type embedding to further enhance SUTrack's performance with minimal overhead. Experiments show that SUTrack outperforms previous task-specific counterparts across 11 datasets spanning five SOT tasks. Moreover, we provide a range of models catering edge devices as well as high-performance GPUs, striking a good trade-off between speed and accuracy. We hope SUTrack could serve as a strong foundation for further compelling research into unified tracking models. Code and models are available at \url{github.com/chenxin-dlut/SUTrack}.

\end{abstract}

\section{Introduction}

Single object tracking (SOT) is a fundamental task in computer vision, focusing on locating an arbitrary target within a video sequence, starting from its initial location. 
Over the years, to broaden the application scenarios of SOT~\cite{SiameseRPN, DiMP, Ocean, transt, TMT, ostrack, artrack, odtrack}, numerous downstream SOT tasks incorporating auxiliary input modalities have been proposed. These tasks include RGB-Depth~\cite{rgbd1k, depthtrack}, RGB-Thermal~\cite{lasher, rgbt234}, RGB-Event~\cite{visevent, COESOT}, and RGB-Language~\cite{TNL2K, TNLS} tracking. 
Existing SOT methods are characterized by fragmentation, with most approaches focusing on one or a few specific downstream tasks and developing separate models for each.

This fragmentation enables customized designs for each task, making it a prevalent choice. However, several deficiencies persist: First, each task requires training a separate model, resulting in redundant parameters and inefficient use of resources.
Second, models are trained on task-specific datasets, which hinders the sharing of knowledge across all available datasets and increases the risk of overfitting.
Third, technological innovations are often repeatedly designed and validated across different tasks, leading to duplicated efforts. 
Although some approaches to unify SOT tasks have emerged, their level of unification remains limited. For instance, some approaches~\cite{vipt,OneTracker,SDSTrack} unify only the architectural design, not the model parameters, while others~\cite{untrack} address only a subset of tasks. This naturally raises the question: \textit{Can a unified visual model address mainstream SOT tasks?}

\begin{figure}[t]
\begin{center}
\includegraphics[width=1\linewidth]{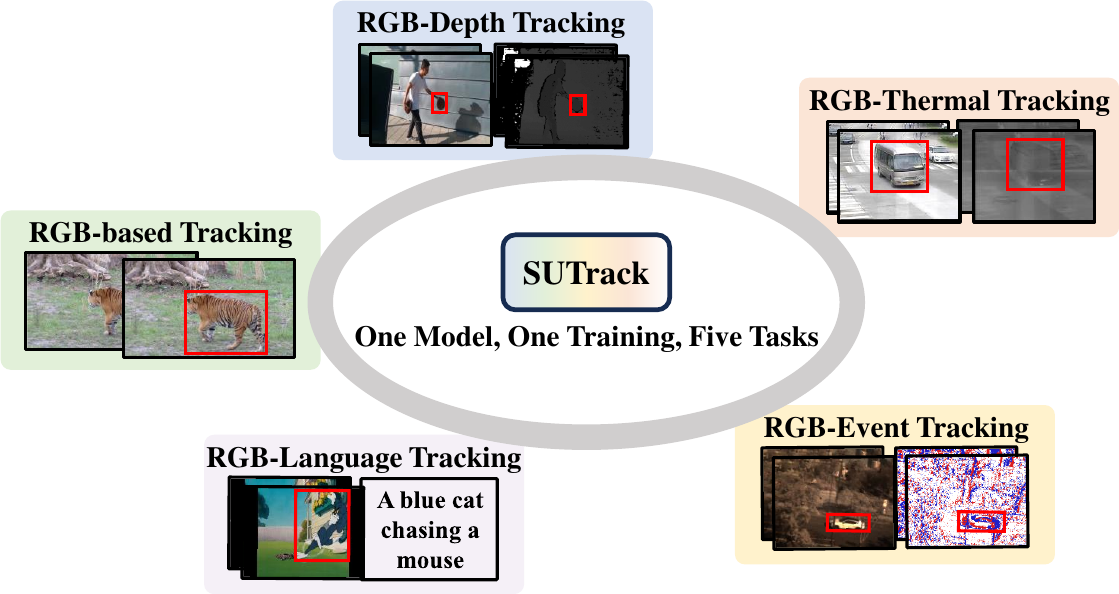}
\end{center}
   \caption{Our SUTrack unifies five SOT tasks into one model with one training session.} 
\label{fig:pipeline}
\end{figure}

To explore this question, we propose a simple and unified framework for SOT, named SUTrack. 
SUTrack unifies five mainstream SOT tasks: RGB-based, RGB-Depth, RGB-Thermal, RGB-Event, and RGB-Language tracking. 
It is based on a straightforward one-stream tracking architecture~\cite{ostrack, mixformer, simtrack}.
By making concise improvements to the interface to accommodate various modalities, 
SUTrack achieves unification with a single model and a single training session. The underlying intuition is that modern general visual models should inherently be capable of integrating knowledge from different modalities. 
We simply need to convert these modalities into a unified form to train the model, rather than developing separate models for each modality.

To this end, we convert the RGB, depth, thermal, event, and language modalities into a unified token format for input into the vision transformer. Specifically, the depth, thermal, and event modalities are typically paired with the RGB modality in image format. Therefore, we modify the patch embedding layer of the vision transformer from three channels to six channels to accommodate channel-concatenated RGB-Depth, RGB-Thermal, or RGB-Event image pairs. 
These image pairs are converted into token embeddings by the modified patch embedding layer and can then be directly fed into the transformer. 
Unlike prevalent methods that employ additional branches to receive auxiliary modalities, this approach is more efficient, adding only $0.06$ M parameters and less than $0.7$ GFlops compared to a purely RGB-based tracker. For the language modality, we employ a CLIP~\cite{clip} text encoder to convert the language input into a token embedding. 
We adopt a vision transformer to process these tokens, followed by a common center-based tracking head~\cite{ostrack} to predict the result. 

Additionally, we introduce a task-recognition auxiliary training strategy. Alongside standard tracking supervision, this approach involves classifying the source task of the input data during training. 
We found that incorporating this task-specific information enhances performance. Importantly, this strategy is used only during training and does not add any overhead during inference.
Furthermore, the cropped template and search region can potentially cause confusion regarding token types (template background, template foreground, and search region)~\cite{LoRAT}, especially for depth, thermal, and event data, which are typically less detailed than RGB data. To address this issue, We develop a soft token type embedding, drawing inspiration from the token type embedding introduced in LoRAT~\cite{LoRAT}. This enhancement equips the model with more precise token type information.

Experiments demonstrate that our SUTrack method is effective, achieving new state-of-the-art performance across 11 benchmarks and five SOT tasks. For instance, SUTrack-B384 attains 74.4\% AUC on the RGB-based benchmark LaSOT, surpassing the recent ODTrack-B384~\cite{odtrack} by 1.2\% while maintaining a similar model size. Moreover, when compared to recent multi-modal trackers~\cite{OneTracker, SDSTrack}, SUTrack consistently outperforms them across all evaluated datasets. It is worth noting that all these prior methods either train different models for each task or cannot cover all five SOT tasks, whereas our SUTrack handles all tasks with a unified model.

In summary, the contributions of this work are two-fold:
\begin{itemize}[leftmargin=0.468cm]
    \item{We propose a simple yet unified SOT framework. It consolidates five SOT tasks into a unified model and learning paradigm. We believe this achievement will significantly reduce the research complexity across SOT tasks.}
    \item{We present a new family of unified tracking models that strike a good balance between speed and accuracy. Experiments confirm the effectiveness of these new models.}
\end{itemize}

\section{Related Work}
\subsection{RGB-based Object Tracking} 
RGB-based object tracking refers to SOT using only RGB data, typically serving as the foundation for downstream SOT tasks. RGB-based object tracking has witnessed significant progress~\cite{SINT,SiameseFC,SiamRPNplusplus,SiamFC++,keeptrack,ToMP,sbt,SLT,CSWinTT,AiATrack,swintrack} over the years, driven by advancements in deep models~\cite{AlexNet,ResNet,DETR}.

Recently, one-stream transformer-based trackers~\cite{mixformer,ostrack,simtrack} have initiated a new revolution in RGB-based object tracking. 
This framework more thoroughly utilizes the capabilities of pretrained transformers by jointly performing feature extraction and fusion, achieving new leading performance.
Building on these pioneering works, we advance further in this paper by developing a new one-stream unified tracking framework through simple modifications to the input interface and training strategy. Our framework not only handles RGB-based object tracking tasks effectively but also performs multi-modal downstream SOT tasks simultaneously, showcasing the greater potential of the one-stream framework combined with modern pretrained transformer models.

\begin{figure*}[t]
\begin{center}
\includegraphics[width=\linewidth]{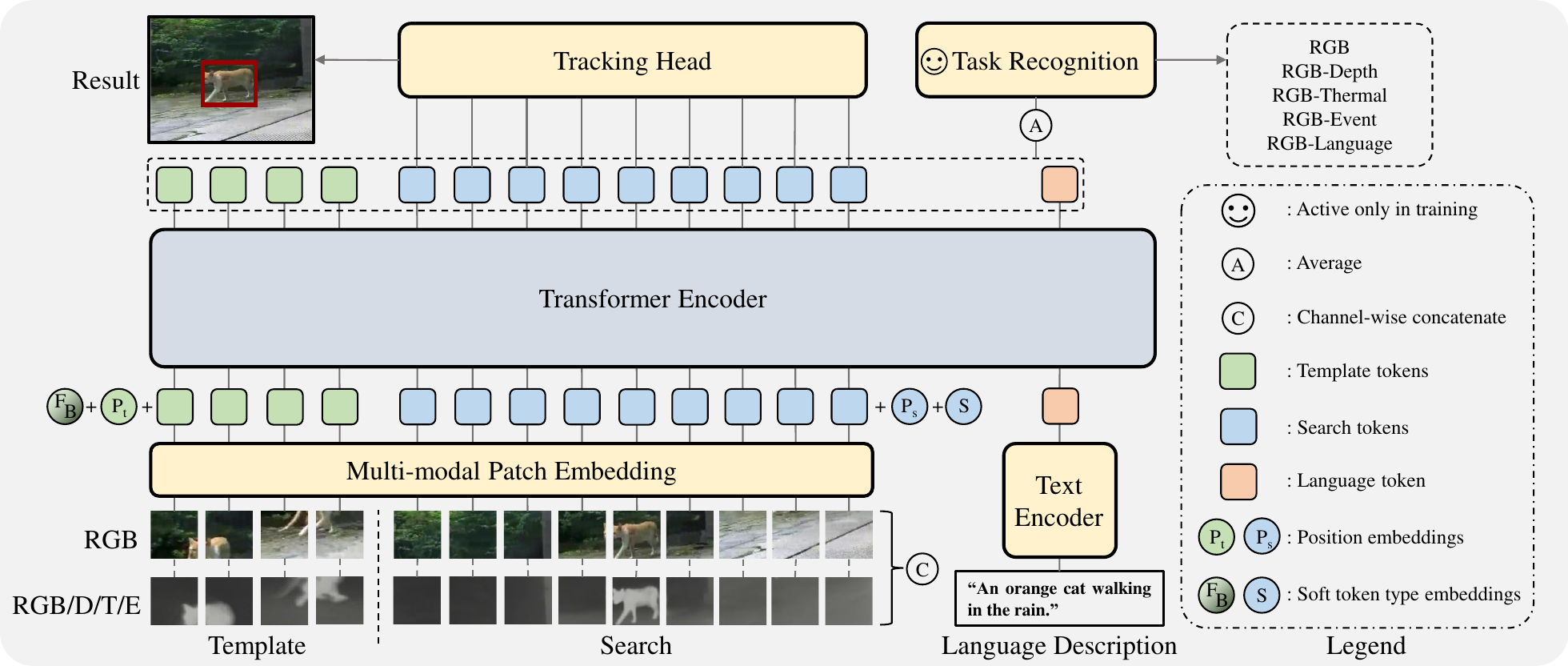}
\end{center}
   \caption{Architecture of the proposed SUTrack. SUTrack unifies five SOT tasks (RGB-based, RGB-Depth, RGB-Thermal, RGB-Event, RGB-Language Tracking) into a single model. We use a unified token embedding format to represent different modalities and train a transformer-based tracking model with these embeddings. In the figure, D/T/E denote depth, thermal, and event modalities, respectively.} 
\label{fig:framework}
\end{figure*}

\subsection{Multi-Modal Object Tracking}
To address the challenges faced by RGB-based tracking in complex or specific scenarios, multi-modal tracking tasks and methods~\cite{ca3dms, jmmac, SNLT, protrack} have been proposed. These tasks integrate auxiliary modalities beyond the RGB input, expanding the applicability of tracking algorithms. Common multi-modal tracking tasks now include RGB-Depth, RGB-Thermal, RGB-Event, and RGB-Language tracking. By incorporating depth~\cite{depthtrack}, thermal~\cite{lasher}, event~\cite{visevent}, or language~\cite{TNL2K} information, multi-modal trackers significantly enhance their ability to tackle issues such as occlusions, low lighting, extreme weather, and target variations.

Despite their impressive performance, existing multi-modal methods typically rely on modality-specific designs and training, \emph{i.e.}, developing different models for each modality. This situation leads to inefficient use of data, computational resources, and human effort. 
In contrast, our approach integrates all multi-modal tracking tasks into a single, unified model and training paradigm. With just one training session, this unified model efficiently handles multiple multi-modal tracking tasks and achieves new state-of-the-art performance across these tasks.

\subsection{Unified Object Tracking Models}
With the advancement of foundational models~\cite{2017Attention,ViT,swintransformer}, it has become feasible to use unified frameworks or models~\cite{pix2seqv2,SAM,flamingo,ofa,uninext,SEGGPT} to address multiple tasks. 
Recently, several works have emerged that aim to unify multiple SOT tasks. ViPT~\cite{vipt} addresses three multi-modal tasks (RGB-Depth, RGB-Thermal, and RGB-Event) within a unified framework using prompt learning, but it does not achieve model-level unification. Subsequently, SDSTrack~\cite{SDSTrack} achieves framework-level unification, while un-track~\cite{untrack} realizes model-level unification for these three tasks. 
OneTracker~\cite{OneTracker} unifies more tasks within a unified framework but does not achieve model-level unification.
Despite these significant strides towards SOT unification, their level of unification remains incomplete. 
In this paper, our SUTrack, for the first time, integrates all five common SOT tasks into a single, streamlined model, further advancing the unification of SOT tasks.

\section{SUTrack}

The overall framework of SUTrack is illustrated in Fig.~\ref{fig:framework}. It adopts a streamlined one-stream transformer architecture.
First, input data from various modalities (including RGB, depth, thermal, event, and natural language) are converted into a unified embedding form. This unified representation enables the model to be trained to handle multiple SOT tasks. Next, positional embeddings and the proposed soft token type embeddings are added to the unified embeddings, enhancing positional information and providing precise prior knowledge about the token type (background/foreground).
The vision transformer encoder then processes and associates these embeddings jointly. The resulting feature embeddings are used to support the final predictions, which are implemented using a center-based tracking head~\cite{ostrack}. Additionally, we introduce a task-recognition prediction, used exclusively during training, to help the model better differentiate between tasks.

\subsection{Unified Modality Representation}
To enable the one-stream transformer model to handle various SOT tasks, we convert the different modality inputs of each task into a unified token embedding form. 

For the RGB-Depth, RGB-Thermal, and RGB-Event tracking tasks, the RGB data are paired with auxiliary modality data (depth, thermal, and event modalities, collectively referred to as DTE). Instead of converting the RGB and DTE modalities into separate token embeddings, we bind them together and jointly convert them using a proposed multi-modal patch embedding. This approach does not add significant computational overhead to the subsequent network. Specifically, the RGB image $\mathbf{I}_{\text{RGB}} \in {\mathbb{R}}^{ H \times W \times 3}$  and the DTE image $\mathbf{I}_{\text{DTE}} \in {\mathbb{R}}^{ H \times W \times 3}$ (noting that DTE data is stored as 3-channel images in current tracking datasets) are concatenated along the channel dimension, resulting in the concatenated image $\mathbf{I}_{\text{concat}} \in {\mathbb{R}}^{ H \times W \times 6}$, summarized as follows:
\begin{equation}
\begin{split}
   \mathbf{I}_{\text{concat}} = \begin{bmatrix}
   \mathbf{I}_{\text{RGB}} \\
   \mathbf{I}_{\text{DTE}}
   \end{bmatrix}.
\end{split}
\end{equation}
Next, $\mathbf{I}_{\text{concat}}$ is divided into fixed-size patches, each with dimensions $P$$\times$$P$$\times$$6$, where $P$ is the patch size. Each $P$$\times$$P$$\times$$6$ patch is then flattened into a one-dimensional vector of size $6P^2$. Finally, a linear transformation is applied to map the flattened patch vectors into an embedding space, as described by the following equation:
\begin{equation}
\begin{split}
   \mathbf{E}^{(i)} = \mathbf{W}_p \mathbf{P}^{(i)} + \mathbf{b}_p,
\end{split}
\end{equation}
 where $\mathbf{E}^{(i)}$ represents the embedding vector of the $i$-th patch with dimension $D$, $\mathbf{P}^{(i)}$ denotes the flattened vector of the $i$-th patch, $\mathbf{W}_p$ is the weight matrix of dimensions $D$$\times$$6P^2$, and $\mathbf{b}_p$ is the bias term with dimension $D$. In this manner, the RGB-DTE data is transformed into a unified token embedding representation.
 For SOT tasks that do not include DTE data, such as RGB-based and RGB-Language tracking, we also use this multi-modal patch embedding by duplicating the RGB channels to create a 6-channel input. 
 
For the language modality in RGB-Language tracking, we use a language model (CLIP-L~\cite{clip} with an additional linear layer to adjust dimensions in our implementation) as the text encoder to extract a single-token feature embedding. This embedding is then concatenated with the multi-modal embeddings and fed into the transformer. For SOT tasks that do not include the language modality, we substitute with a fixed, nonsensical sentence.

\subsection{Soft Token Type Embedding}

LoRAT~\cite{LoRAT} proposes using token type embeddings to explicitly annotate type information for token embeddings, which enhances the distinction between the template foreground, template background, and the search region. However, token embeddings at the edges of the target bounding box often contain both foreground and background information, making it inaccurate to classify them solely as either template foreground or template background. To address this issue, this work introduces a soft token type embedding method to effectively account for both types in these cases.

Specifically, given a template image $\mathbf{I}^{t}_{\text{concat}} \in \mathbb{R}^{H \times W \times 6}$ with a bounding box $\mathbf{B}$ around the target, we first create a mask $\mathbf{M} \in \mathbb{R}^{H \times W}$ of the same size as the image. In this mask, the pixels inside the bounding box are assigned a value of 1, while pixels outside the bounding box are assigned a value of 0:
\begin{equation}
\begin{split}
\mathbf{M}(i, j) = 
\begin{cases} 
1 & \text{if } (i, j) \text{ is inside } \mathbf{B}, \\
0 & \text{otherwise}.
\end{cases}
\end{split}
\end{equation}
Next, we divide the mask $\mathbf{M}$ into non-overlapping patches of size $P \times P$. The $k$-th patch is denoted as $\mathbf{M}_{\text{patch}}^{(k)}$. For each patch, we compute the average value:
\begin{equation}
\begin{split}
\mathbf{m}_{\text{avg}}^{(k)} = \frac{1}{P^2} \sum_{(i,j) \in \mathbf{M}_{\text{patch}}^{(k)}} \mathbf{M}(i, j),
\end{split}
\end{equation}
where $\mathbf{m}_{\text{avg}}^{(k)}$ represents the average value of the $k$-th patch, indicating the degree to which the patch is considered foreground. Based on this average value, we enhance the multi-modal patch embeddings of the image with the corresponding token type embeddings. Specifically, for the \(k\)-th multi-modal patch embedding, we adjust it as follows:
\begin{equation}
\mathbf{E}_{\text{adj}}^{(k)} = \mathbf{E}^{(k)} + \mathbf{m}_{\text{avg}}^{(k)} \cdot \mathbf{E}_{\text{fg}} + (1 - \mathbf{m}_{\text{avg}}^{(k)}) \cdot \mathbf{E}_{\text{bg}},
\end{equation}
where \(\mathbf{E}_{\text{adj}}^{(k)}\) denotes the adjusted embedding for the \(k\)-th patch, \(\mathbf{E}^{(k)}\) represents the original multi-modal patch embedding, \(\mathbf{E}_{\text{fg}}\) is the foreground token type embedding, and \(\mathbf{E}_{\text{bg}}\) is the background token type embedding. This adjustment supplements the embeddings with more accurate foreground and background type information.

For the search region, where bounding box information is not available, we simply add a search region token type embedding \(\mathbf{E}_{\text{search}}\) to each multi-modal patch embedding:
\begin{equation}
\mathbf{E}_{\text{adj}}^{(k)} = \mathbf{E}^{(k)} + \mathbf{E}_{\text{search}}.
\end{equation}

\subsection{Task-recognition Training Strategy}
To enhance the model's ability to differentiate between various tasks, we introduce a task-recognition auxiliary training strategy. This approach explicitly teaches the model to identify the current task. 
Specifically, we compute the average of all feature embeddings output by the transformer model, resulting in a single vector $\mathbf{E}_{\text{avg}}$:
\begin{equation}
\mathbf{E}_{\text{avg}} = \frac{1}{N} \sum_{i=1}^{N} \mathbf{E}^{(i)}_{\text{output}}, 
\end{equation}
where $N$ denotes the number of output token embeddings, and $\mathbf{E}^{(i)}_{\text{output}}$ represents the $i$-th output token embedding. This vector is then processed by a three-layer perceptron to classify the five tasks: RGB-based, RGB-Depth, RGB-Thermal, RGB-Event, and RGB-Language tracking:
\begin{equation}
\mathbf{y}_{\text{task}} = \text{MLP}(\mathbf{E}_{\text{avg}}),
\end{equation}
where \(\text{MLP}(\cdot)\) represents the three-layer perceptron used for task classification, and the output \(\mathbf{y}_{\text{task}}\) represents the predicted probabilities for the five tasks.
The predicted task probabilities \(\mathbf{y}_{\text{task}}\) are then used to compute the cross-entropy loss against the true task labels \(\mathbf{y}_{\text{true}}\):
\begin{equation}
\mathcal{L}_{\text{task}} = -\sum_{j=1}^{K} \mathbf{y}^{(j)}_{\text{true}} \log(\mathbf{y}^{(j)}_{\text{task}}),
\end{equation}
where \(K\) denotes the number of tasks (5 in our case), \(\mathbf{y}^{(j)}_{\text{true}}\) represents the ground truth label for task \(j\), and \(\mathbf{y}^{(j)}_{\text{task}}\) denotes the predicted probability for task \(j\). Experimental results (see Section ``Ablation and Analysis") demonstrate that this explicit task supervision enhances the model's performance. We note that this task-recognition strategy is used exclusively during training and does not impact the inference process.

\subsection{Training and Inference}
We train the model by mixing data from all five SOT tasks in each batch. This strategy allows the model to handle all five tasks after a single training phase. For tracking predictions, following OSTrack~\cite{ostrack}, we use a weighted focal loss~\cite{cornernet} for classification and a combination of $\ell_1$ loss and generalized IoU~\cite{GIoU} loss for regression. For task-recognition predictions, we use the cross-entropy loss as described earlier. The overall loss function is summarized as:
\begin{equation}
\mathcal{L} = \mathcal{L}_{\text{class}} + \lambda_G \mathcal{L}_{\text{IoU}} + \lambda_{L_1} \mathcal{L}_{L_1} + \mathcal{L}_{\text{task}},
\end{equation}
where $\mathcal{L}_{\text{class}}$ denotes the weighted focal loss used for classification, $\mathcal{L}_{\text{IoU}}$ represents the generalized IoU loss, $\mathcal{L}_{L_1}$ is the $\ell_1$ regression loss, $\mathcal{L}_{\text{task}}$ is the cross-entropy loss for task-recognition, and $\lambda_G = 2$ and $\lambda_{L_1} = 5$ are the regularization parameters. 
For inference, we adopt a conventional template update strategy~\cite{Stark}, employing two templates: one static and one updated dynamically during tracking. The template update mechanism is governed by a straightforward approach, utilizing a fixed interval and a confidence threshold to determine when updates occur.

\begin{table}[t]
\centering
\caption{Details of SUTrack model variants.}
\label{tab-model}
\setlength{\tabcolsep}{0.5mm}{
\small
\scalebox{0.77}{
\begin{tabular}{l| c c c r c c}
\toprule
\multirow{2}{*}{Model}           & ~Transformer~ & ~Search~ &~Template~ & \multicolumn{1}{c}{~Params~} & ~FLOPs~ & ~Speed~ \\
~ &~Encoder~ & ~Resolution~ & ~Resolution~  & \multicolumn{1}{c}{(M)} & (G) & (\emph{fps}) \\
\midrule 
SUTrack-L384~~   & HiViT-L       &   $384$$\times$$384$ &$192$$\times$$192$   &$247$$_($$_+$$_{85}$$_)$  &$223$  &12  \\
SUTrack-L224   & HiViT-L       &   $224$$\times$$224$ &$112$$\times$$112$  &$247$$_($$_+$$_{85}$$_)$ &$76$ &35  \\
SUTrack-B384   & HiViT-B       &   $384$$\times$$384$  &$192$$\times$$192$ &$70$$_($$_+$$_{85}$$_)$  &$67$  &32  \\
SUTrack-B224   & HiViT-B     &   $224$$\times$$224$ &$112$$\times$$112$   &$70$$_($$_+$$_{85}$$_)$  &$23$  &55  \\
SUTrack-T224   & HiViT-T      &   $224$$\times$$224$ &$112$$\times$$112$   &$22$$_($$_+$$_{85}$$_)$  &$6$ &100  \\
\bottomrule
\end{tabular}}
}
\end{table}

\section{Experiments}

\subsection{Implementation Details}
The SUTrack models are implemented using Python 3.8 and PyTorch 1.11. Training is conducted on 4 NVIDIA A40 GPUs, while inference speed is evaluated on a single NVIDIA 2080TI GPU.

\textit{Model.} 
We develop five variants of SUTrack models to strike a trade-off between speed and accuracy, each utilizing different transformer encoders and input resolutions, as detailed in Tab.~\ref{tab-model}.
We adopt HiViT-L~\cite{hivit} as the transformer encoder for SUTrack-L384 and L224, HiViT-B for SUTrack-B384 and B224, and HiViT-T for SUTrack-T224. 
The transformer encoders are initialized with the Fast-iTPN~\cite{fastitpn} pre-trained parameters. 
In addition, we present the model parameters, FLOPs, and inference speed in Tab.~\ref{tab-model}. For the parameters, a subscript of $+85$ denotes those specific to the text encoder CLIP-L, which can be omitted for tasks that do not involve language processing. More details of our models are provided in \emph{appendix}.

\textit{Training.}
Our training data comprises commonly used datasets for five SOT tasks, including COCO~\cite{COCO}, LaSOT~\cite{LaSOT}, GOT-10k~\cite{GOT10K}, TrackingNet~\cite{trackingnet}, VASTTrack~\cite{vasttrack}, DepthTrack~\cite{depthtrack}, VisEvent~\cite{visevent}, LasHeR~\cite{lasher}, and TNL2K~\cite{TNL2K}. In each batch, we sample and mix data from these datasets, with RGB data being sampled at twice the rate of multi-modal data. 
The template and search images are generated by expanding the target bounding boxes by factors of $2$ and $4$, respectively. 
We train the model with AdamW~\cite{AdamW} optimizer. The model is trained for a total of $180$ epochs, with $100,000$ image pairs per epoch. More details are provided in \emph{appendix}.

\textit{Inference.}
The online template update interval is set to $25$, with an update confidence threshold of $0.7$ by default. A Hanning window penalty is applied to incorporate positional prior information in tracking, following standard practices~\cite{transt, ostrack}.

\subsection{State-of-the-Art Comparisons}
We compare our SUTrack with state-of-the-art (SOTA) trackers across 11 benchmarks spanning five tasks: RGB-based, RGB-Depth, RGB-Thermal, RGB-Event, and RGB-Language tracking. We note that SUTrack unifies these SOT tasks within a single model. In contrast, other approaches involve training separate models for each task or addressing only a subset of these tasks. The methods compared in this section are generally the latest high-performance approaches. A more comprehensive comparison with earlier methods is available in the \emph{appendix}. 
Additionally, while SeqTrackv2~\cite{seqtrackv2} is contemporaneous with this work, we include it in the comparison tables for reference but do not directly compare it in the main text.

\begin{table*}[t]\Huge
  \caption{State-of-the-art comparisons on four large-scale benchmarks. Methods employing the large model and the base model are compared separately. The number in the method name represents the resolution of the search region. The top two results are highlight with \textbf{bold} and \underline{underlined} fonts, respectively.}
  \label{tab-sota}
\resizebox{1\linewidth}{!}{
  \setlength{\tabcolsep}{2mm}{  
  \small
  \begin{tabular}{l| ccc c ccc c ccc c ccc}
    \toprule
    \multirow{2}*{Method} & \multicolumn{3}{c}{LaSOT}&& \multicolumn{3}{c}{LaSOT$_{ext}$} && \multicolumn{3}{c}{TrackingNet} && \multicolumn{3}{c}{GOT-10k}\\
    \cline{2-4}
    \cline{6-8}
    \cline{10-12}
    \cline{14-16}
    & AUC&P$_{Norm}$&P && AUC&P$_{Norm}$&P && AUC&P$_{Norm}$&P && AO&SR$_{0.5}$&SR$_{0.75}$\\
    \midrule[0.5pt]
             SUTrack-B384	&\textbf{74.4}	&\textbf{83.9}	&\textbf{81.9} & &\underline{52.9} &63.6 &60.1 & &\textbf{86.5}	&\textbf{90.7}	&\textbf{86.8} & &\textbf{79.3} &\textbf{88.0}	&\textbf{80.0}  \\
             SUTrack-B224	&\underline{73.2}	&\underline{83.4}	&80.5 & &\textbf{53.1} &\underline{64.2} &\underline{60.5} & &\underline{85.7}	&\underline{90.3} &\underline{85.1} & &\underline{77.9} &87.5	&\underline{78.5}  \\
    \midrule[0.1pt]
    ODTrack-B384~\cite{odtrack}	&\underline{73.2}	&83.2	&\underline{80.6} & &52.4 &63.9 &60.1 & &85.1	&90.1	&84.9 & &77.0 &\underline{87.9}	&75.1 \\
    LoRAT-B378~\cite{LoRAT}&72.9&81.9&79.1&     &\textbf{53.1}&\textbf{64.8}&\textbf{60.6}&     &84.2&88.4&83.0&            &73.7&82.6&72.9\\
    ARTrackV2-256~\cite{artrackv2}	&71.6	&80.2	&77.2 & &50.8 &61.9 &57.7 & &84.9	&89.3	&84.5 & &75.9 &85.4	&72.7 \\
    AQATrack-256~\cite{aqatrack} &71.4&81.9&78.6&   &51.2&62.2&58.9&    &83.8&88.6&83.1&      &73.8&83.2&72.1\\
    OneTracker-384~\cite{OneTracker}	&70.5	&79.9	&76.5 & &- &- &- & &83.7	&88.4	&82.7 & &- &-	&- \\ 
    EVPTrack-224~\cite{evptrack} &70.4&80.9&77.2&     &48.7&59.5&55.1&   &83.5&88.3&- &          &73.3&83.6&70.7\\
    MixViT-288~\cite{mixformer_journal}	&69.6	&79.9	&75.9 & &- &- &- &  &83.5 &88.3	&83.5  &  &72.5	&82.4	&69.9\\
    DropTrack-224~\cite{DropMAE}	&71.8	&81.8	&78.1 & &52.7 &63.9 &60.2 &  &- &-	&-  &  &75.9	&86.8	&72.0\\
    ROMTrack-384~\cite{ROMTrack}	&71.4	&81.4	&78.2 & &51.3 &62.4 &58.6 &  &84.1 &89.0	&83.7  &  &74.2	&84.3	&72.4\\
    VideoTrack-256~\cite{videotrack}	&70.2	&-	&76.4 & &- &- &- & &83.8	&88.7	&83.1 & &72.9 &81.9	&69.8 \\
    CiteTracker-384~\cite{CiteTrack}	&69.7	&78.6	&75.7 & &- &- &- &  &84.5 &89.0	&84.2  &  &74.7	&84.3	&73.0\\
    \bottomrule
    \multicolumn{16}{l}{\textit{Trackers with larger models}}\\
        \toprule
             SUTrack-L384	&\textbf{75.2}	&\textbf{84.9}	&\textbf{83.2} & &53.6 &64.2 &60.5 & &\textbf{87.7}	&\textbf{91.7}	&\textbf{88.7} & &\textbf{81.5} &\textbf{89.5}	&\textbf{83.3} \\
             SUTrack-L224	&73.5	&83.3	&80.9 & &\underline{54.0} &65.3 &\underline{61.7} & &\underline{86.5}	&90.9 &\underline{86.7} & &\underline{81.0} &\underline{90.4}	&\underline{82.4} \\
    \midrule[0.1pt] 
    LoRAT-L378~\cite{LoRAT}&\underline{75.1} &84.1&82.0&     &\textbf{56.6}&\textbf{69.0}&\textbf{65.1}&     &85.6&89.7&85.4&            &77.5&86.2&78.1\\
    ODTrack-L384~\cite{odtrack}	&74.0	&\underline{84.2}	&\underline{82.3} & &53.9 &\underline{65.4} &\underline{61.7} & &86.1	&\underline{91.0}	&\underline{86.7} & &78.2 &87.2	&77.3 \\
    ARTrackV2-L384~\cite{artrackv2}	&73.6	&82.8	&81.1 & &53.4 &63.7 &60.2 & &86.1	&90.4	&86.2 & &79.5 &87.8	&79.6 \\   
    ARTrack-L384~\cite{artrack} &73.1&82.2&80.3&    &52.8&62.9&59.7&     &85.6&89.6&86.0&    &78.5&87.4&77.8\\
    MixViT-L384~\cite{mixformer_journal}	&72.4	&82.2	&80.1 & &- &- &- &  &85.4 &90.2	&85.7  &  &75.7	&85.3	&75.1\\
    SeqTrack-L384~\cite{seqtrack} &72.5&81.5&79.3&    &50.7&61.6&57.5&    &85.5&89.8&85.8&    &74.8&81.9&72.2\\
    GRM-L320~\cite{grm}	&71.4	&81.2	&77.9 & &- &- &- &  &84.4 &88.9	&84.0  &  &-	&-	&-\\
    TATrack-L384~\cite{TATrack}	&71.1	&79.1	&76.1 & &- &- &- &  &85.0 &89.3	&84.5  &  &-	&-	&-\\
    CTTrack-L320~\cite{CTTrack}	&69.8	&79.7	&76.2 & &- &- &- &  &84.9 &89.1	&83.5  &  &72.8	&81.3	&71.5\\
  \bottomrule
\end{tabular}
}}
\end{table*}

\begin{table*}
  \centering
  \caption{State-of-the-art comparisons of efficient tracking on four large-scale benchmarks.}
  \label{tab-sota-efficient}
\resizebox{0.98\linewidth}{!}{
  \setlength{\tabcolsep}{2mm}{  
  \small
  \begin{tabular}{l|c ccc c ccc c ccc c ccc c cc}
    \toprule
    \multirow{2}*{Method} &\multicolumn{3}{c}{LaSOT} & &\multicolumn{3}{c}{LaSOT$_{ext}$} & &\multicolumn{3}{c}{TrackingNet} & &\multicolumn{3}{c}{GOT-10k} & &\multicolumn{2}{c}{Speed (\textit{fps})} \\
    \cline{2-4}
    \cline{6-8}
    \cline{10-12}
    \cline{14-16}
    \cline{18-19}
   & AUC&P$_{Norm}$&P & & AUC&P$_{Norm}$&P  & & AUC&P$_{Norm}$&P & & AO&SR$_{0.5}$&SR$_{0.75}$ &&CPU& AGX\\
    \midrule[0.5pt]
    SUTrack-T224 &\textbf{69.6}&\textbf{79.3}&\textbf{75.4}  & &\textbf{50.2}&\textbf{61.1}&\textbf{57.0} & &\textbf{82.7}&\textbf{87.2}&\textbf{80.8} & &\textbf{72.7}&\textbf{82.1}&\textbf{70.5} & &23&34\\
    \midrule[0.1pt]
    MixformerV2-S~\cite{mixformerv2}&60.6&69.9&60.4  & &\underline{43.6}&-&\underline{46.2} & &75.8&81.1&70.4 & &-&-&- & &30&-\\
    HiT~\cite{HiT} &\underline{64.6}&\underline{73.3}&\underline{68.1} & &-&-&- & &\underline{80.0}&\underline{84.4}&\underline{77.3} & &64.0&72.1&\underline{58.1} &&33&61\\
    FEAR-L~\cite{fear} & 57.9&-&60.9 & &-&-&- & &-&-&- & &  64.5&74.6&- && -&-\\
    FEAR-XS~\cite{fear}& 53.5&-&54.5 & &-&-&- & & -&-&- & &  61.9&72.2&- &&60&38\\
    HCAT~\cite{HCAT} & 59.3&68.7&61.0 & &-&-&- & & 76.6&82.6&72.9  & &   \underline{65.1}&\underline{76.5}&56.7 &&45&55 \\
    E.T.Track~\cite{ETTrack}& 59.1&-&- & &-&-&- & & 75.0&80.3&70.6 &&  -&-&- &&47&20\\
    LightTrack~\cite{lighttrack}& 53.8&-&53.7 & &-&-&- & & 72.5&77.8&69.5 & &  61.1&71.0&- &&41&36\\
    ATOM~\cite{ATOM}& 51.5&57.6&50.5 & &-&-&- & & 70.3&77.1&64.8 & &  55.6&63.4&40.2 &&18&22 \\
    ECO~\cite{ECO}& 32.4&33.8&30.1 & &-&-&- & & 55.4&61.8&49.2 &&  31.6&30.9&11.1 &&15&39 \\
  \bottomrule
\end{tabular}
}}
\end{table*}

\begin{table}[t]\normalsize
    \caption{SOTA comparisons on RGB-Depth tracking.}
\label{tab-sota-rgbd}
  \centering
\resizebox{1\linewidth}{!}{
  \setlength{\tabcolsep}{1.25mm}{
    \small
    \begin{tabular}{l|ccc c ccc}
    \toprule
    \multirow{2}*{Method} & \multicolumn{3}{c}{VOT-RGBD22} & & \multicolumn{3}{c}{DepthTrack} \\
        \cline{2-4} \cline{6-8}
 & EAO & Acc. & Rob. & &F-score &Re &Pr \\
    \midrule[0.5pt]
    SUTrack-L384 &\textbf{76.6}&\underline{83.5}&\textbf{92.2} & &\textbf{66.4}&\textbf{66.4}&\textbf{66.5}\\
    SUTrack-L224 &76.4&83.4&\underline{91.9} & &64.3&64.6&64.0\\
    SUTrack-B384 &\textbf{76.6}&\textbf{83.9}&91.4 & &64.4&64.2&\underline{64.6}\\
    SUTrack-B224 &\underline{76.5}&82.8&91.8 & &\underline{65.1}&\underline{65.7}&64.5\\
    SUTrack-T224 &68.1&81.0&83.9 & &61.7&62.1&61.2\\
    \midrule[0.1pt]
    SeqTrackv2-L384~\cite{seqtrackv2} &74.8&82.6&91.0 & &62.3&62.6&62.5\\
    SeqTrackv2-B256~\cite{seqtrackv2} &74.4&81.5&91.0 & &63.2&63.4&62.9 \\
    OneTracker~\cite{OneTracker} &72.7&81.9&87.2 & &60.9&60.4&60.7\\
    SDSTrack~\cite{SDSTrack} &72.8&81.2&88.3 & &61.9&60.9&61.4\\
    Un-Track~\cite{untrack} &72.1&82.0 &86.9 & &61.0&60.8&61.1\\
    ViPT~\cite{vipt} &72.1&81.5 &87.1 & &59.4&59.6&59.2\\
    ProTrack~\cite{protrack} &65.1&80.1&80.2 & &57.8&57.3&58.3\\
    SPT~\cite{rgbd1k} &65.1&79.8&85.1 & &53.8&54.9&52.7\\
    DeT~\cite{depthtrack} &65.7&76.0&84.5 & &53.2&50.6&56.0\\
    DAL~\cite{dal} &-&-&- & &42.9&36.9&51.2\\
    \bottomrule
    \end{tabular}
    }
  }
\end{table}

\textit{RGB-based Tracking.}
We evaluate our SUTrack on four large-scale RGB-based tracking benchmarks, including the long-term benchmarks LaSOT~\cite{LaSOT} and LaSOT$_{ext}$~\cite{lasot_journal}, as well as the short-term benchmarks TrackingNet~\cite{trackingnet} and GOT-10k~\cite{GOT10K}. The results are presented in Tab.~\ref{tab-sota}. Compared to trackers using the base model, our SUTrack-B224 surpasses all previous trackers across all four benchmarks, with the higher resolution SUTrack-B384 delivering even better results. Specifically, SUTrack-B384 achieves AUC scores of 74.4\% on LaSOT, 86.5\% on TrackingNet, and an AO score of 79.3\% on GOT-10k, surpassing the previous best tracker, ODTrack~\cite{odtrack}, by 1.2\%, 1.4\%, and 2.3\% points, respectively. On LaSOT$_{ext}$, SUTrack-B224 matches the performance of the previous best tracker, LoRAT-B378, achieving an AUC score of 53.1\%.
When compared to trackers using large models, SUTrack-L384 and L224 also demonstrate competitive performance, setting new SOTA results on LaSOT, TrackingNet, and GOT-10k, while achieving the second-best performance on LaSOT$_{ext}$.

\textit{Efficient RGB-based Tracking.}
We develop the SUTrack-T224 model for edge devices with limited computational resources, and compare its performance with SOTA efficient trackers. 
The results are detailed in Tab.~\ref{tab-sota-efficient}, which also includes the running speeds on both the Intel Core i9-9900K @ 3.60GHz CPU and the NVIDIA Jetson AGX Xavier edge device. Our method not only achieves real-time speeds on edge devices (with the real-time line defined as 20 \textit{fps} by the VOT challenge~\cite{vot2020}) but also significantly outperforms previous efficient trackers. Specifically, SUTrack-T224 surpasses the previous best performances by 5\%, 6.6\%, 2.7\%, and 7.6\% on LaSOT, LaSOT$_{ext}$, TrackingNet, and GOT-10K, respectively.

\textit{RGB-Depth Tracking.}
For the RGB-Depth tracking task, our SUTrack models set new state-of-the-art performance on both the VOT-RGBD22~\cite{vot2022} and DepthTrack~\cite{depthtrack} benchmarks. Specifically, on the VOT-RGBD22 benchmark, both SUTrack-L384 and SUTrack-B384 achieve an EAO score of 76.6\%, surpassing the previous best, SDSTrack~\cite{SDSTrack}, by 3.8\%. On the DepthTrack benchmark, SUTrack-L384 achieves an F-score of 66.4\%, outperforming SDSTrack by 4.5\%.

\textit{RGB-Thermal Tracking.}
On the LasHeR~\cite{lasher} benchmark, both SUTrack-L384 and SUTrack-L224 achieve an AUC score of 61.9\%, surpassing the previous best, OneTracker, by 8.1\% points. On the RGBT234~\cite{rgbt234} benchmark, SUTrack-L224 obtains an AUC score of 70.8\%, exceeding the performance of OneTracker by 6.6\% points. These results highlight the significant performance advantage of our SUTrack model for RGB-Thermal tracking.

\textit{RGB-Event Tracking.}
SUTrack-L224, L384, and B384 secure the top three positions on the RGB-Event tracking benchmark, VisEvent~\cite{visevent}. Notably, SUTrack-L224 attains the highest AUC score of 64.0\%, surpassing the previous best, OneTracker, by 3.2 \% points.

\textit{RGB-Language Tracking.}
Our five SUTrack models take the top five spots on the RGB-Language tracking benchmark, TNL2K~\cite{TNL2K}. SUTrack-L384 achieves the highest AUC score of 67.9\%, significantly surpassing the previous best, OneTracker, by 9.9 \% points. On the small-scale OTB99~\cite{TNLS} benchmark, SUTrack demonstrates competitive performance despite not using the OTB99 training set, which other algorithms rely on.

\subsection{Ablation and Analysis.}
\label{sec:ablation}
The results of the ablation study are presented in Tab.\ref{tab-ablation}, where SUTrack-B224 serves as the baseline model, as shown in row \#1.

\textit{Multi-Task v.s. Single-Task.}
In Tab.\ref{tab-ablation} (\#2), we train separate models for each SOT task. Compared to our multi-task unified model, single-task models show inferior performance across all tasks. The decline is particularly notable for RGB-Depth, RGB-Thermal, and RGB-Event tracking, where the training data is relatively small. 
This underscores the benefit of multi-task unification, which leverages shared knowledge across tasks to boost overall performance.

\textit{Zero-Shot Performance.}
In Tab.~\ref{tab-ablation} (\#3), we evaluate the zero-shot performance of the model by training it on all tasks' data except for the task being assessed. Although the results indicate a significant drop in performance, the model exhibits some zero-shot generalization capabilities. 
Notably, for specific tasks such as RGB-Depth and RGB-Thermal, the performance is comparable to or even exceeds that of the single-task models reported in \#2.

\textit{Task-recognition Training Strategy.}
Tab.~\ref{tab-ablation} (\#4) presents the results after omitting the task-recognition auxiliary training strategy. 
This results in a decrease in performance compared to our default method. The potential reason for this is that explicit task supervision helps the model differentiate between data types, enabling it to better learn the specific characteristics of each task.

\begin{table}[t]\normalsize
    \caption{SOTA comparisons on RGB-Thermal tracking.}
\label{tab-sota-rgbt}
  \centering
\resizebox{1\linewidth}{!}{
  \setlength{\tabcolsep}{3mm}{
    \small
    \begin{tabular}{l|ccccc}
    \toprule
    \multirow{2}*{Method} & \multicolumn{2}{c}{LasHeR} & & \multicolumn{2}{c}{RGBT234} \\
        \cline{2-3} \cline{5-6}
 & AUC & P & &MSR &MPR \\
    \midrule[0.5pt]
    SUTrack-L384 &\textbf{61.9}&\underline{76.9} & &\underline{70.3}&\underline{93.7}\\
    SUTrack-L224 &\textbf{61.9}&\textbf{77.0} & &\textbf{70.8}&\textbf{94.6}\\
    SUTrack-B384 &60.9&75.8 & &69.2&92.1\\
    SUTrack-B224 &59.9&74.5 & &69.5&92.2\\
    SUTrack-T224 &53.9&66.7 & &63.8&85.9\\
    \midrule[0.1pt]
        SeqTrackv2-L384~\cite{seqtrackv2} &\underline{61.0}&76.7 & &68.0&91.3\\
    SeqTrackv2-B256~\cite{seqtrackv2} &55.8&70.4 & &64.7&88.0 \\
    OneTracker~\cite{OneTracker} &53.8&67.2 & &64.2&85.7\\
    SDSTrack~\cite{SDSTrack} &53.1&66.5 & &62.5&84.8\\
    Un-Track~\cite{untrack} &-&- & &62.5&84.2\\
    ViPT~\cite{vipt} &52.5&65.1 & &61.7&83.5\\
    ProTrack~\cite{protrack} &42.0&53.8 & &59.9&79.5\\
    APFNet~\cite{apfnet} &36.2&50.0 & &57.9&82.7\\
    JMMAC~\cite{jmmac} &-&- & &57.3&79.0\\
    CMPP~\cite{cmpp} &-&- & &57.5&82.3\\
    CAT~\cite{cat} &31.4&45.0 & &56.1&80.4\\
    HMFT~\cite{vtuav} &31.3&43.6 & &-&-\\
    MaCNet~\cite{macnet} &-&- & &55.4&79.0\\
    FANet~\cite{fanet} &30.9&44.1 & &55.3&78.7\\
    DAFNet~\cite{dafnet} &-&- & &54.4&79.6 \\
    \bottomrule
    \end{tabular}
    }
  }
\end{table}

\textit{Data Ratio.}
In multi-task joint training, we sample RGB data at twice the rate of multi-modal data. The results of uniform sampling, as shown in Tab.~\ref{tab-ablation} (\#5), reveal a drop in performance. 
This is due to the limited diversity of multi-modal datasets, where an excessive proportion of such data can hinder model robustness.
We look forward to the availability of larger-scale multi-modal datasets in the future.

\begin{table}[t]\normalsize
    \caption{SOTA comparisons on RGB-Event tracking.}
\label{tab-sota-rgbe}
  \centering
\resizebox{1\linewidth}{!}{
  \setlength{\tabcolsep}{6mm}{
    \small
    \begin{tabular}{l|cc}
    \toprule
    \multirow{2}*{Method} & \multicolumn{2}{c}{VisEvent}\\
        \cline{2-3}
 & AUC & P \\
    \midrule[0.5pt]
    SUTrack-L384 &\underline{63.8}&\underline{80.5} \\
    SUTrack-L224 &\textbf{64.0}&\textbf{80.9} \\
    SUTrack-B384 &63.4&79.8 \\
    SUTrack-B224 &62.7&79.9 \\
    SUTrack-T224 &58.8&75.7 \\
     \midrule[0.1pt]
         SeqTrackv2-L384~\cite{seqtrackv2} &63.4&80.0\\
    SeqTrackv2-B256~\cite{seqtrackv2} &61.2&78.2 \\
    OneTracker~\cite{OneTracker} &60.8&76.7\\
    SDSTrack~\cite{SDSTrack} &59.7 &76.7\\
    Un-Track~\cite{untrack} &58.9 &75.5\\
    ViPT~\cite{vipt} &59.2&75.8\\
    ProTrack~\cite{protrack} &47.1&63.2\\
    OSTrack\_E~\cite{ostrack} &53.4&69.5\\
    SiamRCNN\_E~\cite{SiamRCNN} &49.9&65.9 \\
    TransT\_E~\cite{transt} &47.4&65.0\\
    \bottomrule
    \end{tabular}
    }
  }
\end{table}

\begin{table}\footnotesize
    \caption{SOTA comparisons on RGB-Language tracking.}
\label{tab-sota-rgbl}
  \centering
\resizebox{\linewidth}{!}{
  \setlength{\tabcolsep}{3mm}{  
  \small
  \begin{tabular}{l| cc c cc}
    \toprule
    \multirow{2}*{Method} & \multicolumn{2}{c}{TNL2K} & & \multicolumn{2}{c}{OTB99}\\
    \cline{2-3}
    \cline{5-6}
    & AUC&P && AUC&P\\
    \midrule[0.5pt]
    SUTrack-L384 &\textbf{67.9}&\textbf{72.1} & &71.2&93.1 \\
    SUTrack-L224 &\underline{66.7}&\underline{70.3} & &\underline{72.7}&\underline{94.4}\\
    SUTrack-B384 &65.6&69.3 & &69.7&91.2\\
    SUTrack-B224 &65.0&67.9 & &70.8&93.4\\
    SUTrack-T224 &60.9&62.3 & &67.4&88.6\\
    \midrule[0.1pt]
    SeqTrackv2-L384~\cite{seqtrackv2} &62.4&66.1 & &71.4&93.6\\
SeqTrackv2-B256~\cite{seqtrackv2} &57.5&59.7 & &71.2&93.9 \\
OneTracker~\cite{OneTracker} &58.0&59.1 & &69.7&91.5\\
JointNLT~\cite{JointNLT}	&56.9 &58.1 & &65.3&85.6 \\
DecoupleTNL\cite{DecoupleTNL}&56.7&56.0 & &\textbf{73.8}&\textbf{94.8}\\
Zhao \emph{et al.}~\cite{zhao2023transformervision-languagetracking}	&56.0&- & &69.9&91.2 \\
Li \emph{et al.}\cite{CTRTNL} &44.0&45.0 & &69.0&91.0\\
TNL2K-2~\cite{TNL2K}	&42.0&42.0 & &68.0&88.0 \\
SNLT~\cite{SNLT}	&27.6&41.9 & &66.6&80.4 \\
TransVG~\cite{transvg} &26.1&28.9 & &-&- \\
Feng \emph{et al.}~\cite{feng2019robust}	&25.0&27.0 & &67.0&73.0\\
RTTNLD~\cite{RTTNLD}	&25.0&27.0 & &61.0&79.0 \\
  \bottomrule
\end{tabular}
}}
\end{table}

\begin{table}[t]
\centering
\caption{Ablation Study. $\Delta$ denotes the performance change (averaged over benchmarks) compared with the baseline.
}
\label{tab-ablation}
\small
\resizebox{1\linewidth}{!}{
\setlength{\tabcolsep}{0.5mm}{
\begin{tabular}{l|c|ccccc|c}
\toprule
\# &Method &LaSOT &VOT-RGBD22  &LasHeR &VisEvent &TNL2K  &$\Delta$\\
\midrule
1 &Baseline &73.2&76.5&59.9&62.7&65.0 &--\\
2 &Multi-Task $\rightarrow$ Single-Task &72.7&57.0&50.1&56.7&61.8 &\textbf{-7.8}\\  
3 &Multi-Task $\rightarrow$ Zero-Shot &58.2&62.3&49.1&50.1&58.8 &\textbf{-11.8}\\   
4 &W/o Task Recognition &72.6&76.5&59.8&62.5&63.9 &\textbf{-0.4}\\ 
5 &More RGB $\rightarrow$ Uniform &71.6&75.8&59.2&62.0&64.3 &\textbf{-0.9}\\ 
6 &Separate Representation &72.0&78.2&61.2&65.2&65.2 &\textbf{+0.9}\\  
7 &Concat $\rightarrow$ Mul &63.8&58.0&48.6&54.0&54.0 &\textbf{-11.8}\\ 
8 &Concat $\rightarrow$ Add &73.0&76.4&60.0&62.4&64.9 &\textbf{-0.3}\\ 
9 &W/o Token Type Embedding 
&72.4&76.2&59.4&61.5&64.6&\textbf{-0.6}\\ 
10 &Soft $\rightarrow$ Hard 
&72.7&76.3&59.8 &62.4&64.7 &\textbf{-0.3}\\ 
\bottomrule
\end{tabular}
}}
\end{table}

\textit{Depth/Thermal/Event Modality Representation.}
We use multi-modal patch embedding to jointly represent RGB and Depth/Thermal/Event (DTE) modality image pairs. 
In Tab.~\ref{tab-ablation} (\#6), we explore an alternative, more computationally intensive method: applying standard patch embedding separately to RGB and DTE modalities, and then concatenating them along the spatial dimension.
This approach yields higher performance, highlighting the potential of SUTrack. 
However, it results in nearly double the computational load. For efficiency, we have chosen to use our default multi-modal patch embedding method.

\textit{Language Modality Combination.}
In Tab.~\ref{tab-ablation} (\#7 and \#8), we investigate two alternative methods for combining language features with image features: one through multiplication and the other through addition. Both methods result in lower performance compared to our default approach.

\textit{Token Type Embedding.}
In Tab.~\ref{tab-ablation} (\#9 and \#10), we compare our soft token type embedding with results from both the absence of token type embedding and the original hard token type embedding method~\cite{LoRAT}. Our soft token type embedding achieves superior performance by providing more precise token type information, which aids the model in distinguishing between template background, foreground, and search region tokens.

More ablation studies are provided in \emph{appendix}.

\section{Conclusion}
This work proposes a simple yet unified SOT framework, \textit{i.e.}, SUTrack, which integrates five SOT tasks into a unified model trained in one session. SUTrack shows that a single model with a unified input representation is capable of managing diverse SOT tasks, eliminating the necessity for separate task-specific models or training processes. Extensive experiments demonstrate that SUTrack is effective, achieving competitive performance across all five SOT tasks. We hope SUTrack could serve as a solid foundation for future research on unified single object tracking.


\appendix

\section{Appendix}

In this appendix, we provide additional content to complement the main manuscript:
\begin{itemize}[leftmargin=0.468cm]
    \item{More implementation details.}
    \item{Introduction of benchmarks.}
    \item{Unification comparison.}
    \item{Additional state-of-the-art comparisons.}
    \item{Additional ablation study.}
    \item{Limitation.}
\end{itemize}

\section{More Implementation Details}
This section provides the implementation details that are omitted from the main manuscript due to space constraints.

\subsection{Devices}
The training of SUTrack are conducted on Intel Xeon Gold 6330 CPU @ 2.00GHz with 512 GB RAM and 4 NVIDIA A40 GPUs with 48GB memory. The speed in Tab.1 of the main manuscript is measured on Intel Core i9-9900K CPU @ 3.60GHz with 64 GB RAM and a single 2080 TI GPU. The speeds reported in Tab. 3 of the main manuscript are measured on an Intel Core i9-9900K @ 3.60GHz CPU and an NVIDIA Jetson AGX Xavier edge device, respectively.

\subsection{Model} 
Here, we provide more details of our SUTrack model.
For the multi-modal patch embedding, the weight matrix \(\mathbf{W}_p\) has dimensions $D$ $\times$$(P^2$$\times$$6)$, whereas the corresponding weight matrix in the pre-trained model~\cite{fastitpn} has dimensions $D$ $\times$$(P^2$$\times$$3)$. To align the pre-trained model’s parameters with \(\mathbf{W}_p\), we first expand the pre-trained parameters by repeating them along the last dimension, resulting in dimensions $D$ $\times$$(P^2$$\times$$6)$. We then divide these expanded parameters by 2 to maintain a numerical range consistent with the original pre-trained model. The adjusted parameters are then loaded into our model. The dimensions 
$D$ varies depending on the encoder model used. For SUTrack-L, SUTrack-B, and SUTrack-T, 
$D$ is set to 768, 512, and 384, respectively. For the text encoder, we use the pre-trained CLIP-L's text encoder~\cite{clip}, augmented with a linear layer to align its dimensions with those of the transformer encoder. The CLIP-L model is kept frozen during training to preserve the knowledge gained from its pre-trained language data.
For SOT tasks that do not include the language modality, we use the padding token from CLIP as a fixed, nonsensical sentence in place of the language description. The architecture of the tracking head follows that of OSTrack~\cite{ostrack}. The task-recognition head is implemented as a three-layer perceptron with a hidden dimension of 256.

\subsection{Training}
Here, we provide more details of the training of SUTrack.
The template and search images are generated by expanding the target bounding boxes by factors of $2$ and $4$, respectively. Data augmentation is performed using horizontal flipping and brightness jittering. We train the model with AdamW~\cite{AdamW} optimizer. The learning rate is set to $1$$e$$-5$ for the transformer, $1$$e-$$4$ for the remaining unfrozen modules, and the weight decay is set to $1$$e$$-4$. The model is trained for a total of $180$ epochs, with $100,000$ image pairs per epoch. The learning rate is reduced by a factor of 10 after 144 epochs.

\begin{table*}[t]
\centering
\caption{Comparison of unification levels and unified tasks.
}
\label{tab-unification}
\small
\resizebox{1\linewidth}{!}{
\setlength{\tabcolsep}{3mm}{
\begin{tabular}{l|cccccccc}
\toprule
\multirow{2}*{Method} &\multicolumn{2}{c}{Unification} & &\multicolumn{5}{c}{Task}\\
\cline{2-3}
\cline{5-9}
&Framework-level &Model-level  & &RGB-based &RGB-Depth &RGB-Thermal  &RGB-Event &RGB-Language\\
\midrule
SUTrack &$\surd$&$\surd$ & &$\surd$&$\surd$&$\surd$&$\surd$&$\surd$\\
\midrule
SeqTrackv2~\cite{seqtrackv2}&$\surd$&$\surd$ & &&$\surd$&$\surd$&$\surd$&$\surd$ \\
OneTracker~\cite{OneTracker}&$\surd$& & &$\surd$&$\surd$&$\surd$&$\surd$&$\surd$ \\
SDSTrack~\cite{SDSTrack} &$\surd$& & &&$\surd$&$\surd$&$\surd$&\\
UnTrack~\cite{untrack} &$\surd$&$\surd$ & &&$\surd$&$\surd$&$\surd$&\\
ViPT~\cite{vipt} &$\surd$& & &&$\surd$&$\surd$&$\surd$&\\ 
ProTrack~\cite{protrack} &$\surd$& & &&$\surd$&$\surd$&$\surd$&\\
\bottomrule
\end{tabular}
}}
\end{table*}

\begin{table}[t]\normalsize
    \caption{SOTA comparisons on RGB-Depth tracking.}
\label{tab-sota-rgbd-supp}
  \centering
\resizebox{1\linewidth}{!}{
  \setlength{\tabcolsep}{1.25mm}{
    \small
    \begin{tabular}{l|ccc c ccc}
    \toprule
    \multirow{2}*{Method} & \multicolumn{3}{c}{VOT-RGBD22} & & \multicolumn{3}{c}{DepthTrack} \\
        \cline{2-4} \cline{6-8}
 & EAO & Acc. & Rob. & &F-score &Re &Pr \\
    \midrule[0.5pt]
    SUTrack-L384 &\textbf{76.6}&\underline{83.5}&\textbf{92.2} & &\textbf{66.4}&\textbf{66.4}&\textbf{66.5}\\
    SUTrack-L224 &76.4&83.4&\underline{91.9} & &64.3&64.6&64.0\\
    SUTrack-B384 &\textbf{76.6}&\textbf{83.9}&91.4 & &64.4&64.2&\underline{64.6}\\
    SUTrack-B224 &\underline{76.5}&82.8&91.8 & &\underline{65.1}&\underline{65.7}&64.5\\
    SUTrack-T224 &68.1&81.0&83.9 & &61.7&62.1&61.2\\
    \midrule[0.1pt]
    SeqTrackv2-L384~\cite{seqtrackv2} &74.8&82.6&91.0 & &62.3&62.6&62.5\\
    SeqTrackv2-B256~\cite{seqtrackv2} &74.4&81.5&91.0 & &63.2&63.4&62.9 \\
    OneTracker~\cite{OneTracker} &72.7&81.9&87.2 & &60.9&60.4&60.7\\
    SDSTrack~\cite{SDSTrack} &72.8&81.2&88.3 & &61.9&60.9&61.4\\
    Un-Track~\cite{untrack} &72.1&82.0 &86.9 & &61.0&60.8&61.1\\
    ViPT~\cite{vipt} &72.1&81.5 &87.1 & &59.4&59.6&59.2\\
    ProTrack~\cite{protrack} &65.1&80.1&80.2 & &57.8&57.3&58.3\\
    SPT~\cite{rgbd1k} &65.1&79.8&85.1 & &53.8&54.9&52.7\\
    SBT-RGBD~\cite{sbt} &70.8&80.9&86.4 & &-&-&-\\
    OSTrack~\cite{ostrack} &67.6&80.3&83.3 & &52.9&52.2&53.6\\
    DeT~\cite{depthtrack} &65.7&76.0&84.5 & &53.2&50.6&56.0\\
    DMTrack~\cite{vot2022} &65.8&75.8&85.1 & &-&-&-\\
    DDiMP~\cite{vot2020} &-&-&- & &48.5&56.9&50.3\\
    STARK-RGBD~\cite{Stark} &64.7&80.3&79.8 & &-&-&-\\
    KeepTrack~\cite{keeptrack} &60.6&75.3&79.7 & &-&-&-\\
    DRefine~\cite{vot2021} &59.2&77.5&76.0 & &-&-&-\\
    DAL~\cite{dal} &-&-&- & &42.9&36.9&51.2\\
    ATCAIS~\cite{vot2020} &55.9&76.1&73.9 & &47.6&45.5&50.0\\
    LTMU-B~\cite{LTMU} &-&-&- & &46.0&41.7&51.2\\
    GLGS-D~\cite{vot2020} &-&-&- & &45.3&36.9&58.4\\
    LTDSEd~\cite{VOT2019} &-&-&- & &40.5&38.2&43.0\\
    Siam-LTD~\cite{vot2020} &-&-&- & &37.6&34.2&41.8\\
    SiamM-Ds~\cite{VOT2019} &-&-&- & &33.6&26.4&46.3\\
    CA3DMS~\cite{ca3dms} &-&-&- & &22.3&22.8&21.8\\
    DiMP~\cite{DiMP} &54.3&70.3&73.1 & &-&-&-\\
    ATOM~\cite{ATOM} &50.5&59.8&68.8 & &-&-&-\\
    \bottomrule
    \end{tabular}
    }
  }
\end{table}

\begin{table}[t]\normalsize
    \caption{SOTA comparisons on RGB-Thermal tracking.}
\label{tab-sota-rgbt-supp}
  \centering
\resizebox{1\linewidth}{!}{
  \setlength{\tabcolsep}{3mm}{
    \small
    \begin{tabular}{l|ccccc}
    \toprule
    \multirow{2}*{Method} & \multicolumn{2}{c}{LasHeR} & & \multicolumn{2}{c}{RGBT234} \\
        \cline{2-3} \cline{5-6}
 & AUC & P & &MSR &MPR \\
    \midrule[0.5pt]
    SUTrack-L384 &\textbf{61.9}&\underline{76.9} & &\underline{70.3}&\underline{93.7}\\
    SUTrack-L224 &\textbf{61.9}&\textbf{77.0} & &\textbf{70.8}&\textbf{94.6}\\
    SUTrack-B384 &60.9&75.8 & &69.2&92.1\\
    SUTrack-B224 &59.9&74.5 & &69.5&92.2\\
    SUTrack-T224 &53.9&66.7 & &63.8&85.9\\
    \midrule[0.1pt]
    SeqTrackv2-L384~\cite{seqtrackv2} &\underline{61.0}&76.7 & &68.0&91.3\\
    SeqTrackv2-B256~\cite{seqtrackv2} &55.8&70.4 & &64.7&88.0 \\
    OneTracker~\cite{OneTracker} &53.8&67.2 & &64.2&85.7\\
    SDSTrack~\cite{SDSTrack} &53.1&66.5 & &62.5&84.8\\
    Un-Track~\cite{untrack} &-&- & &62.5&84.2\\
    ViPT~\cite{vipt} &52.5&65.1 & &61.7&83.5\\
    ProTrack~\cite{protrack} &42.0&53.8 & &59.9&79.5\\
    OSTrack~\cite{ostrack} &41.2&51.5 & &54.9&72.9\\
    TransT~\cite{transt} &39.4&52.4 &&\\
    APFNet~\cite{apfnet} &36.2&50.0 & &57.9&82.7\\
    JMMAC~\cite{jmmac} &-&- & &57.3&79.0\\
    CMPP~\cite{cmpp} &-&- & &57.5&82.3\\
    STARK~\cite{Stark} &36.1&44.9 &&\\
    mfDiMP~\cite{mfdimp} &34.3&44.7 & &42.8&64.6\\
    DAPNet~\cite{dapnet} &31.4&43.1 & &-&-\\
    CAT~\cite{cat} &31.4&45.0 & &56.1&80.4\\
    HMFT~\cite{vtuav} &31.3&43.6 & &-&-\\
    MaCNet~\cite{macnet} &-&- & &55.4&79.0\\
    FANet~\cite{fanet} &30.9&44.1 & &55.3&78.7\\
    DAFNet~\cite{dafnet} &-&- & &54.4&79.6 \\
    SGT~\cite{sgt} &25.1&36.5 & &47.2&72.0 \\
    \bottomrule
    \end{tabular}
    }
  }
\end{table}

\begin{table}[t]\normalsize
    \caption{SOTA comparisons on RGB-Event tracking.}
\label{tab-sota-rgbe-supp}
  \centering
\resizebox{1\linewidth}{!}{
  \setlength{\tabcolsep}{6mm}{
    \small
    \begin{tabular}{l|cc}
    \toprule
    \multirow{2}*{Method} & \multicolumn{2}{c}{VisEvent}\\
        \cline{2-3}
 & AUC & P \\
    \midrule[0.5pt]
    SUTrack-L384 &\underline{63.8}&\underline{80.5} \\
    SUTrack-L224 &\textbf{64.0}&\textbf{80.9} \\
    SUTrack-B384 &63.4&79.8 \\
    SUTrack-B224 &62.7&79.9 \\
    SUTrack-T224 &58.8&75.7 \\
     \midrule[0.1pt]
    SeqTrackv2-L384~\cite{seqtrackv2} &63.4&80.0\\
    SeqTrackv2-B256~\cite{seqtrackv2} &61.2&78.2 \\
    OneTracker~\cite{OneTracker} &60.8&76.7\\
    SDSTrack~\cite{SDSTrack} &59.7 &76.7\\
    Un-Track~\cite{untrack} &58.9 &75.5\\
    ViPT~\cite{vipt} &59.2&75.8\\
    ProTrack~\cite{protrack} &47.1&63.2\\
    OSTrack\_E~\cite{ostrack} &53.4&69.5\\
    SiamRCNN\_E~\cite{SiamRCNN} &49.9&65.9 \\
    TransT\_E~\cite{transt} &47.4&65.0\\
    LTMU\_E~\cite{LTMU} &45.9&65.5 \\
    PrDiMP\_E~\cite{PrDiMP} &45.3&64.4 \\
    STARK\_E~\cite{Stark} &44.6&61.2 \\
    MDNet\_E~\cite{MDNet} &42.6&66.1 \\
    SiamCar\_E~\cite{Stark} &42.0&59.9 \\
    VITAL\_E~\cite{VITAL} &41.5&64.9 \\
    ATOM\_E~\cite{ATOM} &41.2&60.8 \\
    SiamBAN\_E~\cite{SiamBAN} &40.5&59.1 \\
    SiamMask\_E~\cite{SiamMask} &36.9&56.2 \\
    \bottomrule
    \end{tabular}
    }
  }
\end{table}

\begin{table}\footnotesize
    \caption{SOTA comparisons on RGB-Language tracking.}
\label{tab-sota-rgbl-supp}
  \centering
\resizebox{\linewidth}{!}{
  \setlength{\tabcolsep}{3mm}{  
  \small
  \begin{tabular}{l| cc c cc}
    \toprule
    \multirow{2}*{Method} & \multicolumn{2}{c}{TNL2K} & & \multicolumn{2}{c}{OTB99}\\
    \cline{2-3}
    \cline{5-6}
    & AUC&P && AUC&P\\
    \midrule[0.5pt]
    SUTrack-L384 &\textbf{67.9}&\textbf{72.1} & &71.2&93.1 \\
    SUTrack-L224 &\underline{66.7}&\underline{70.3} & &\underline{72.7}&\underline{94.4}\\
    SUTrack-B384 &65.6&69.3 & &69.7&91.2\\
    SUTrack-B224 &65.0&67.9 & &70.8&93.4\\
    SUTrack-T224 &60.9&62.3 & &67.4&88.6\\
    \midrule[0.1pt]
SeqTrackv2-L384~\cite{seqtrackv2} &62.4&66.1 & &71.4&93.6\\
SeqTrackv2-B256~\cite{seqtrackv2} &57.5&59.7 & &71.2&93.9 \\
OneTracker~\cite{OneTracker} &58.0&59.1 & &69.7&91.5\\
JointNLT~\cite{JointNLT}	&56.9 &58.1 & &65.3&85.6 \\
DecoupleTNL\cite{DecoupleTNL}&56.7&56.0 & &\textbf{73.8}&\textbf{94.8}\\
Zhao \emph{et al.}~\cite{zhao2023transformervision-languagetracking}	&56.0&- & &69.9&91.2 \\
CapsuleTNL~\cite{CapsuleTNL}	&- &- & &71.1&92.4\\
Li \emph{et al.}\cite{CTRTNL} &44.0&45.0 & &69.0&91.0\\
TNL2K-2~\cite{TNL2K}	&42.0&42.0 & &68.0&88.0 \\
SNLT~\cite{SNLT}	&27.6&41.9 & &66.6&80.4 \\
GTI~\cite{GTI} &-&- & &58.1&73.2\\
TransVG~\cite{transvg} &26.1&28.9 & &-&- \\
Feng \emph{et al.}~\cite{feng2019robust}	&25.0&27.0 & &67.0&73.0\\
RTTNLD~\cite{RTTNLD}	&25.0&27.0 & &61.0&79.0 \\
Wang \emph{et al.}~\cite{wang2018describe}	&-&-  & &65.8&89.1 \\
TNLS~\cite{TNLS}	&-&-  & &55.3&72.3 \\
OneStage-BERT~\cite{onestagebert} &19.8&-  & &24.6&32.2 \\
LBYL-BERT~\cite{onestagebert} &18.3&-  & &20.7&26.0\\
  \bottomrule
\end{tabular}
}}
\end{table}

\begin{table}[t]\large
    \caption{SOTA comparisons on NFS and UAV123 benchmarks in AUC score.}
\label{tab-sota-small}
  \centering
\resizebox{1\linewidth}{!}{
  \setlength{\tabcolsep}{6mm}{
    \small
    \begin{tabular}{l|cc}
    \toprule
    Method &NFS&UAV123\\
    \midrule[0.5pt]
    SUTrack-L384 &\underline{70.3}&70.6	 \\
    SUTrack-L224 &69.8&70.9	 \\
    SUTrack-B384 &69.3&70.4 \\
    SUTrack-B224 &\textbf{71.3}&\underline{71.7}	 \\
    SUTrack-T224 &68.4&69.4	 \\
    \midrule[0.1pt]
    ARTrackV2-L384~\cite{artrackv2}&68.4&\underline{71.7} \\
    LoRAT-L378~\cite{LoRAT}&66.7&\textbf{72.5}	\\
    ARTrack-L384~\cite{artrack} &67.9 &71.2\\
    SeqTrack-L384~\cite{seqtrack} &66.2 &68.5 \\
    OSTrack~\cite{ostrack}&66.5&70.7\\
    SimTrack~\cite{simtrack}&- &71.2 \\
    STARK~\cite{Stark}&66.2&68.2 \\
    TransT~\cite{transt}&65.7&69.1 \\
    TrDiMP~\cite{TMT}&66.5&67.5 \\
    DiMP~\cite{DiMP}&61.8&64.3 \\
    Ocean~\cite{Ocean}&49.4&57.4 \\
   ATOM~\cite{ATOM}&58.3&63.2 \\
       ECO~\cite{ECO}&52.2&53.5 \\
    RT-MDNet~\cite{RTMDNet}&43.3&52.8 \\
    SiamFC\cite{SiameseFC}&37.7&46.8 \\
    \bottomrule
    \end{tabular}
    }
  }
\end{table}

\section{Introduction of Benchmarks}
In this section, we provide detailed descriptions of the benchmarks used for evaluation.

\subsection{RGB-based Tracking Benchmarks}

\hspace{3.5mm}\textit{LaSOT}. LaSOT~\cite{LaSOT} is a large-scale, long-term tracking dataset. Its test set comprises 280 videos with an average length of 2,448 frames. The evaluated metrics include Success (AUC) and Precision (P and $\rm{P}_{Norms}$) scores, with AUC being the primary metric.

\textit{LaSOT$_{ext}$.} LaSOT$_{ext}$~\cite{lasot_journal} is an extension of the long-term LaSOT dataset. It comprises 150 video sequences across 15 new object classes. The evaluation metrics are consistent with those used for the LaSOT dataset.

\textit{TrackingNet.}
TrackingNet~\cite{trackingnet} is a large-scale short-term tracking dataset that encompasses a wide range of object classes and scenes. The test set consists of 511 sequences. We submit the tracking results of our SUTrack to the official online evaluation server to obtain the Success (AUC) and Precision (P and P$_{Norm}$) scores.

\textit{GOT-10k.}
GOT-10k~\cite{GOT10K} is a large-scale short-term tracking dataset. Its test set consists of 180 videos that cover a broad spectrum of common tracking challenges. We submit the tracking results to the official evaluation server. The evaluated metrics include Average Overlap (AO) and Success Rates (SR$_{0.5}$ and SR$_{0.75}$).

\textit{VOT.} VOT2020~\cite{vot2020} and VOT2022~\cite{vot2022} comprises 60 challenging videos, respectively. They employ an anchor-based evaluation method, running a tracker from numerous starting frames. The primary metric used is expected average overlap (EAO), which simultaneously assesses the tracker's accuracy and robustness. Trackers can submit either mask predictions or bounding box predictions for evaluation.

\textit{NFS.}
The NFS~\cite{NFS} dataset is a small-scale benchmark consisting of 100 challenging videos, primarily featuring fast-moving targets. The main evaluation metric is the Success (AUC) score.

\textit{UAV123.}
UAV123~\cite{UAV} is a small-scale tracking benchmark that includes 123 low-altitude aerial videos. The primary evaluation metric is the Success (AUC) score.

\subsection{RGB-Depth Tracking Benchmarks.}

\hspace{3.5mm}\textit{DepthTrack.}
DepthTrack~\cite{depthtrack} is a comprehensive benchmark for long-term RGB-Depth tracking. The test set consists of 50 videos, each annotated with 15 per-frame attributes. The primary evaluation metric is the F-score, which is commonly used in long-term tracking tasks.

\textit{VOT-RGBD22.}
VOT-RGBD2022~\cite{vot2022} is a recent tracking benchmark consisting of 127 RGB-Depth sequences. The evaluation protocol uses an anchor-based method similar to that employed by VOT2020. The primary performance metric is expected average overlap (EAO).

\begin{table*}[!ht]\Huge
  \caption{State-of-the-art comparisons on four large-scale benchmarks. Methods employing the large model and the base model are compared separately. The top two results are highlight with \textbf{bold} and \underline{underlined} fonts, respectively.}
  \label{tab-sota-supp}
\resizebox{1\linewidth}{!}{
  \setlength{\tabcolsep}{1.5mm}{  
  \small
  \begin{tabular}{l| ccc c ccc c ccc c ccc}
    \toprule
    \multirow{2}*{Method} & \multicolumn{3}{c}{LaSOT}&& \multicolumn{3}{c}{LaSOT$_{ext}$} && \multicolumn{3}{c}{TrackingNet} && \multicolumn{3}{c}{GOT-10k}\\
    \cline{2-4}
    \cline{6-8}
    \cline{10-12}
    \cline{14-16}
    & AUC&P$_{Norm}$&P && AUC&P$_{Norm}$&P && AUC&P$_{Norm}$&P && AO&SR$_{0.5}$&SR$_{0.75}$\\
    \midrule[0.5pt]
             SUTrack-B384	&\textbf{74.4}	&\textbf{83.9}	&\textbf{81.9} & &\underline{52.9} &63.6 &60.1 & &\textbf{86.5}	&\textbf{90.7}	&\textbf{86.8} & &\textbf{79.3} &\textbf{88.0}	&\textbf{80.0}  \\
             SUTrack-B224	&\underline{73.2}	&\underline{83.4}	&80.5 & &\textbf{53.1} &\underline{64.2} &\underline{60.5} & &\underline{85.7}	&\underline{90.3} &85.1 & &\underline{77.9} &87.5	&\underline{78.5}  \\
    \midrule[0.1pt]
    ODTrack-B384~\cite{odtrack}	&\underline{73.2}	&83.2	&\underline{80.6} & &52.4 &63.9 &60.1 & &85.1	&90.1	&84.9 & &77.0 &\underline{87.9}	&75.1 \\
    LoRAT-B378~\cite{LoRAT}&72.9&81.9&79.1&     &\textbf{53.1}&\textbf{64.8}&\textbf{60.6}&     &84.2&88.4&83.0&            &73.7&82.6&72.9\\
    ARTrackV2-384~\cite{artrackv2}	&73.0	&82.0	&79.6 & &\underline{52.9} &63.4 &59.1 & &\underline{85.7}	&89.8	&\underline{85.5} & &77.5 &86.0	&75.5 \\
    AQATrack-256~\cite{aqatrack} &71.4&81.9&78.6&   &51.2&62.2&58.9&    &83.8&88.6&83.1&      &73.8&83.2&72.1\\
    OneTracker-384~\cite{OneTracker}	&70.5	&79.9	&76.5 & &- &- &- & &83.7	&88.4	&82.7 & &- &-	&- \\ 
    EVPTrack-224~\cite{evptrack} &70.4&80.9&77.2&     &48.7&59.5&55.1&   &83.5&88.3&- &          &73.3&83.6&70.7\\
    MixViT-288~\cite{mixformer_journal}	&69.6	&79.9	&75.9 & &- &- &- &  &83.5 &88.3	&83.5  &  &72.5	&82.4	&69.9\\
    DropTrack-224~\cite{DropMAE}	&71.8	&81.8	&78.1 & &52.7 &63.9 &60.2 &  &- &-	&-  &  &75.9	&86.8	&72.0\\
    ROMTrack-384~\cite{ROMTrack}	&71.4	&81.4	&78.2 & &51.3 &62.4 &58.6 &  &84.1 &89.0	&83.7  &  &74.2	&84.3	&72.4\\
    ARTrack-384~\cite{artrack} &72.6&81.7&79.1&    &51.9&62.0&58.5&     &85.1&89.1&84.8&    &75.5&84.3&74.3\\
    VideoTrack-256~\cite{videotrack}	&70.2	&-	&76.4 & &- &- &- & &83.8	&88.7	&83.1 & &72.9 &81.9	&69.8 \\
    SeqTrack-B384~\cite{seqtrack} &71.5&81.1&77.8&    &50.5&61.6&57.5&    &83.9&88.8&83.6&    &74.5&84.3&71.4\\
    GRM-B256~\cite{grm}	&69.9	&79.3	&75.8 & &- &- &- &  &84.0 &88.7	&83.3  &  &73.4	&82.9	&70.4\\
    CiteTracker-384~\cite{CiteTrack}	&69.7	&78.6	&75.7 & &- &- &- &  &84.5 &89.0	&84.2  &  &74.7	&84.3	&73.0\\
    TATrack-B224~\cite{TATrack}	&69.4	&78.2	&74.1 & &- &- &- &  &83.5 &88.3	&81.8  &  &73.0	&83.3	&68.5\\
    CTTrack-B320~\cite{CTTrack}	&67.8	&77.8	&74.0 & &- &- &- &  &82.5 &87.1	&80.3  &  &71.3	&80.7	&70.3\\
    OSTrack-384~\cite{ostrack}	&71.1	&81.1	&77.6 & &50.5 &61.3 &57.6 &  &83.9 &88.5	&83.2  &  &73.7	&83.2	&70.8\\
    SimTrack-B224~\cite{simtrack}	&69.3	&78.5	&-  &  &- &- &-  &  &82.3 &86.5	&-  &  &68.6	&78.9	&62.4\\
    RTS~\cite{RTS}	&69.7	&76.2	&73.7  &  &- &- &-  &  &81.6 &86.0  &79.4  &  &-	&-	&-\\
    SwinTrack~\cite{swintrack}	&71.3	&-	&76.5 & &49.1 &- &55.6 & &84.0	&-	&82.8 &	&72.4	&-	&67.8\\
    Mixformer-22k~\cite{mixformer}	&69.2	&78.7	&74.7 & &- &- &- & &83.1	&88.1	&81.6 & &70.7	&80.0	&67.8\\
    AiATrack~\cite{AiATrack}	&69.0	&79.4	&73.8  &  &47.7 &55.6 &55.4  & &82.7 &87.8  &80.4  &  &69.6	&63.2	&80.0\\
    UTT~\cite{UTT}	&64.6	&-	&67.2 & &- &- &- & &79.7 &- &77.0 & &67.2	&76.3	&60.5\\
    CSWinTT~\cite{CSWinTT}	&66.2	&75.2	&70.9 & &- &- &- & &81.9	&86.7	&79.5 &	&69.4	&78.9	&65.4\\
    SLT~\cite{SLT}	&66.8	&75.5	&- & &- &- &- & &82.8 &87.5	&81.4 & &67.5	&76.5	&60.3\\
    SBT~\cite{sbt}	&66.7	&-	&71.1 & &- &- &- & &-	&-	&-	& &70.4	&80.8	&64.7\\
    ToMP~\cite{ToMP}	&68.5	&79.2	&73.5 & &45.9 &- &- & &81.5	&86.4	&78.9 &	&-	&-	&-\\
    KeepTrack~\cite{keeptrack}	&67.1	&77.2	&70.2 & &48.2 &- &- & &-	&-	&-	& &-	&-	&-\\
    STARK~\cite{Stark}	&67.1	&77.0	&- & &- &- &- & &82.0	&86.9	&- &	&68.8	&78.1	&64.1\\
    TransT~\cite{transt}	&64.9	&73.8	&69.0 & &- &- &- & &81.4	&86.7	&80.3 &	&67.1	&76.8	&60.9\\
    TrDiMP~\cite{TMT}  	&63.9	&-	&61.4 & &- &- &- &	&78.4	&83.3 	&73.1	&	&68.8	&80.5	&59.7\\
    AutoMatch~\cite{automatch} 	&58.3	&-	&59.9 & &- &- &- & &76.0	&-	&72.6	& &65.2	&76.6	&54.3\\
    DSTrpn~\cite{DSTrpn}	&43.4	&54.4	&- & &- &- &-	& &64.9	&-	&58.9 &	&-	&-	&-\\
    SiamAttn~\cite{SiamAtt}  	&56.0	&64.8	&- & &- &- &- &	&75.2	&81.7	&-	&	&-	&-	&-\\
    SiamBAN~\cite{SiamBAN}  	&51.4	&59.8	&- & &- &- &- &	&-	&- 	&-	&	&-	&-	&-\\
    Ocean~\cite{Ocean}	&56.0	&65.1	&56.6 & &- &- &-	& &-	&-	&- &	&61.1	&72.1	&47.3\\
    SiamR-CNN~\cite{SiamRCNN}  	&64.8	&72.2	&- &	&- &- &- & &81.2	&85.4	&80.0	&	&64.9	&72.8	&59.7\\
    DiMP~\cite{DiMP}	   	&56.9	&65.0	&56.7 &	&39.2 &47.6 &45.1 & &74.0	&80.1	&68.7 &	&61.1	&71.7	&49.2\\
    SiamPRN++~\cite{SiamRPNplusplus}	&49.6	&56.9	&49.1 & &34.0 &41.6 &39.6 &	&73.3	&80.0	&69.4 &	&51.7	&61.6	&32.5\\
    ATOM~\cite{ATOM}	   	&51.5	&57.6	&50.5 &	&37.6 &45.9 &43.0 & &70.3	&77.1	&64.8	& &55.6	&63.4	&40.2\\
    MDNet~\cite{MDNet}	   	 &39.7	&46.0	&37.3	& &27.9 &34.9 &31.8 & &60.6	&70.5	&56.5	& &29.9	&30.3	&9.9\\
    \bottomrule
    \multicolumn{16}{l}{\textit{Trackers with larger models}}\\
        \toprule
             SUTrack-L384	&\textbf{75.2}	&\textbf{84.9}	&\textbf{83.2} & &53.6 &64.2 &60.5 & &\textbf{87.7}	&\textbf{91.7}	&\textbf{88.7} & &\textbf{81.5} &\textbf{89.5}	&\textbf{83.3} \\
             SUTrack-L224	&73.5	&83.3	&80.9 & &\underline{54.0} &65.3 &\underline{61.7} & &\underline{86.5}	&90.9 &\underline{86.7} & &\underline{81.0} &\underline{90.4}	&\underline{82.4} \\
    \midrule[0.1pt] 
    LoRAT-L378~\cite{LoRAT}&\underline{75.1} &84.1&82.0&     &\textbf{56.6}&\textbf{69.0}&\textbf{65.1}&     &85.6&89.7&85.4&            &77.5&86.2&78.1\\
    ODTrack-L384~\cite{odtrack}	&74.0	&\underline{84.2}	&\underline{82.3} & &53.9 &\underline{65.4} &\underline{61.7} & &86.1	&\underline{91.0}	&\underline{86.7} & &78.2 &87.2	&77.3 \\
    ARTrackV2-L384~\cite{artrackv2}	&73.6	&82.8	&81.1 & &53.4 &63.7 &60.2 & &86.1	&90.4	&86.2 & &79.5 &87.8	&79.6 \\   
    ARTrack-L384~\cite{artrack} &73.1&82.2&80.3&    &52.8&62.9&59.7&     &85.6&89.6&86.0&    &78.5&87.4&77.8\\
    MixViT-L384~\cite{mixformer_journal}	&72.4	&82.2	&80.1 & &- &- &- &  &85.4 &90.2	&85.7  &  &75.7	&85.3	&75.1\\
    SeqTrack-L384~\cite{seqtrack} &72.5&81.5&79.3&    &50.7&61.6&57.5&    &85.5&89.8&85.8&    &74.8&81.9&72.2\\
    GRM-L320~\cite{grm}	&71.4	&81.2	&77.9 & &- &- &- &  &84.4 &88.9	&84.0  &  &-	&-	&-\\
    TATrack-L384~\cite{TATrack}	&71.1	&79.1	&76.1 & &- &- &- &  &85.0 &89.3	&84.5  &  &-	&-	&-\\
    CTTrack-L320~\cite{CTTrack}	&69.8	&79.7	&76.2 & &- &- &- &  &84.9 &89.1	&83.5  &  &72.8	&81.3	&71.5\\
    UNINEXT-L~\cite{uninext}	&72.4	&80.7	&78.9 & &54.4 &61.8 &61.4 & &85.1	&88.2	&84.7 & &- &-	&- \\
    Mixformer-L320~\cite{mixformer}	&70.1	&79.9	&76.3 & &- &- &- & &83.9 &88.9	&83.1 & &-	&-	&-\\
    Unicorn~\cite{unicorn}	&68.5	&-	&-  &  &- &- &-  &  &83.0 &86.4	&82.2  &  &-	&-	&-\\
    SimTrack-L224~\cite{simtrack}	&70.5	&79.7	&-  &  &- &- &-  &  &83.4 &87.4	&-  &  &69.8	&78.8	&66.0\\
  \bottomrule
\end{tabular}
}}
\end{table*}

\subsection{RGB-Thermal Tracking Benchmarks}

\hspace{3.5mm}\textit{LasHeR.}
LasHeR~\cite{lasher} is a highly diverse and comprehensive benchmark for RGB-Thermal tracking. 
Its test set contains 245 testing video sequences. The evaluated metrics include AUC and Precision (P) scores.

\textit{RGBT234.}
RGBT234~\cite{rgbt234} is a substantial RGB-Thermal tracking benchmark featuring 234 videos that include visible and thermal infrared pairs. The evaluated metrics include Maximum Success Rate(MSR) and Maximum Precision Rate(MPR) scores.

\subsection{RGB-Event Tracking Benchmarks}

\hspace{3.5mm}\textit{VisEvent.}
VisEvent~\cite{visevent} is an extensive RGB-Event tracking benchmark based on data collected from real-world environments. Its test set comprises 320 videos, and the evaluated metrics include AUC and Precision (P) scores.

\subsection{RGB-Language Tracking benchmarks.}

\hspace{3.5mm}\textit{TNL2K.}
TNL2K~\cite{TNL2K} is a large-scale benchmark for RGB-language tracking. The test set contains 700 videos, each accompanied by language annotations and bounding box annotations to indicate the target. The performance is evaluated using AUC and Precision (P) scores.

\textit{OTB99.}
OTB99~\cite{TNLS} is a small-scale benchmark for RGB-language tracking, derived by supplementing language annotations to the OTB100 dataset. The evaluation metrics include AUC and Precision (P) scores.

\section{Unification Comparison}
As discussed in the main manuscript, while some approaches~\cite{protrack,vipt,untrack,SDSTrack,OneTracker,seqtrackv2} have emerged to unify SOT tasks, their level of unification remains limited. These methods either train separate models for each task or address only a subset of the tasks. In Tab.~\ref{tab-unification}, we compare the unification levels and the range of tasks covered by our SUTrack method with those of other methods. Framework-level unification refers to unifying the framework across various tasks, while model-level unification involves unifying the model parameters as well. 
Our SUTrack is the only method that achieves both framework-level and model-level unification while supporting all five SOT tasks.

\section{Additional State-of-the-Art Comparisons}
In this section, we provide additional state-of-the-art (SOTA) comparisons, including more compared methods and benchmarks.

\subsection{Comparisons with more methods}
In the ``State-of-the-Art Comparisons" section of the main manuscript, the methods compared are generally the latest high-performance approaches. Here, we provide more comprehensive comparisons including earlier methods in Tab.~\ref{tab-sota-rgbd-supp}, Tab.~\ref{tab-sota-rgbt-supp}, Tab.~\ref{tab-sota-rgbe-supp}, 
Tab.~\ref{tab-sota-rgbl-supp}, 
and Tab.~\ref{tab-sota-supp}. These additional results further validate the effectiveness of our method, which continues to achieve state-of-the-art performance.

\begin{figure}[t!]
\centering
\begin{center}
    \includegraphics[width=1\linewidth]{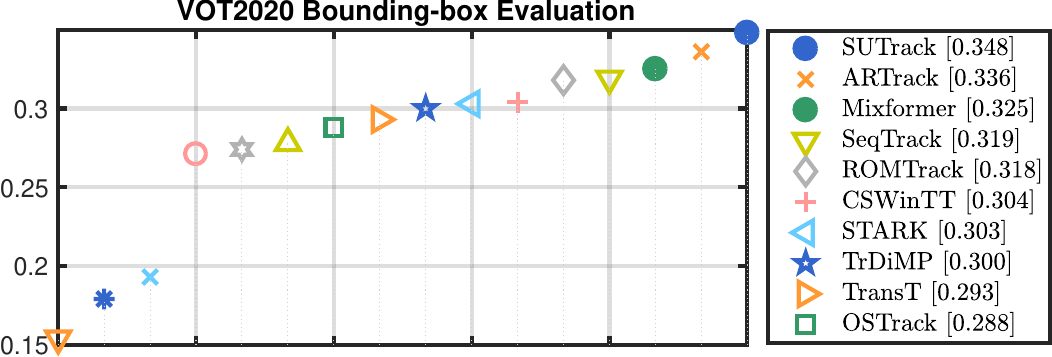}
\end{center}
\begin{center}
\includegraphics[width=1\linewidth]{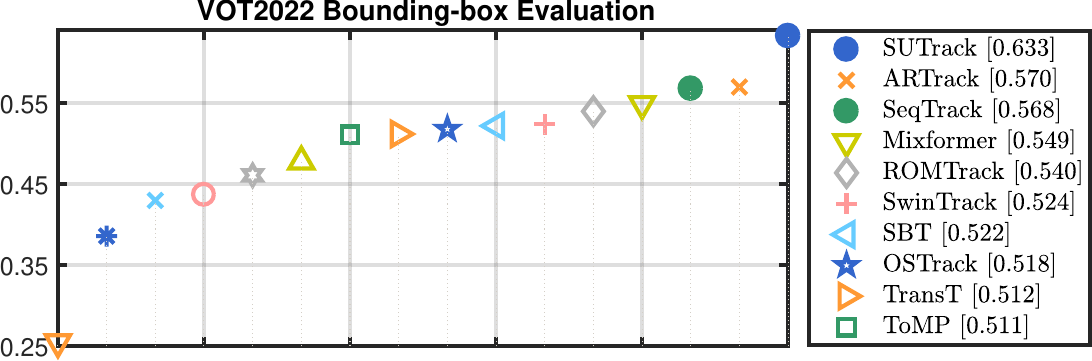}
\end{center}
\caption{EAO rank plots on VOT2020 and VOT2022.}
\label{fig:votrank}
\end{figure}

\begin{table}[t]
\centering
\caption{Ablation Study. $\Delta$ denotes the performance change (averaged over benchmarks) compared with the baseline.
}
\label{tab-ablation-supp}
\small
\resizebox{1\linewidth}{!}{
\setlength{\tabcolsep}{0.5mm}{
\begin{tabular}{l|c|ccccc|c}
\toprule
\# &Method &LaSOT &VOT-RGBD22  &LasHeR &VisEvent &TNL2K  &$\Delta$\\
\midrule
1 &Baseline &73.2&76.5&59.9&62.7&65.0 &--\\
2 &Multi-Modal $\rightarrow$ RGB-only &--&74.9&51.5&58.4&64.6 &\textbf{-3.7}\\  
3 &W/o Task Recognition &72.6&76.5&59.8&62.5&63.9 &\textbf{-0.4}\\ 
4 &Text Token &72.9&76.1&59.8&62.6&64.8 &\textbf{-0.2}\\ 
5 &Task Token &73.0&76.7&60.0&62.4&64.9 &\textbf{-0.1}\\ 
6 &Half-Copy $\rightarrow$ Full-Copy &72.9&75.8&60.1&62.1&64.7 &\textbf{-0.3}\\ 
7 &Half-Copy $\rightarrow$ Single-Copy &73.2&77.3&59.2&62.1&64.7 &\textbf{-0.2}\\ 
\bottomrule
\end{tabular}
}}
\end{table}

\subsection{Comparisons on NFS and UAV123}
We provide SOTA comparisons on two additional small-scale benchmarks: NFS~\cite{NFS} and UAV123~\cite{UAV}. Tab.~\ref{tab-sota-small} shows that our SUTrack models also achieve competitive results, matching or surpassing the most recent trackers, ARTrackV2~\cite{artrackv2} and LORAT~\cite{LoRAT}.

\subsection{Comparisons on VOT2020 and VOT2022}
We evaluate our models on VOT2020~\cite{vot2020} and VOT2022~\cite{vot2022} benchmarks by submitting the bounding boxes using their protocol. The compared methods are also based on bounding box predictions. For methods with multiple models, we report the performance of their best-performing official model. As shown in Fig.~\ref{fig:votrank}, our SUTrack achieves the highest EAO scores of 34.8\% and 63.3\% on VOT2020 and VOT2022, respectively.

\subsection{Attribute-based Comparisons on LaSOT}

\begin{figure}[t!]
\centering
    \includegraphics[width=1\linewidth]{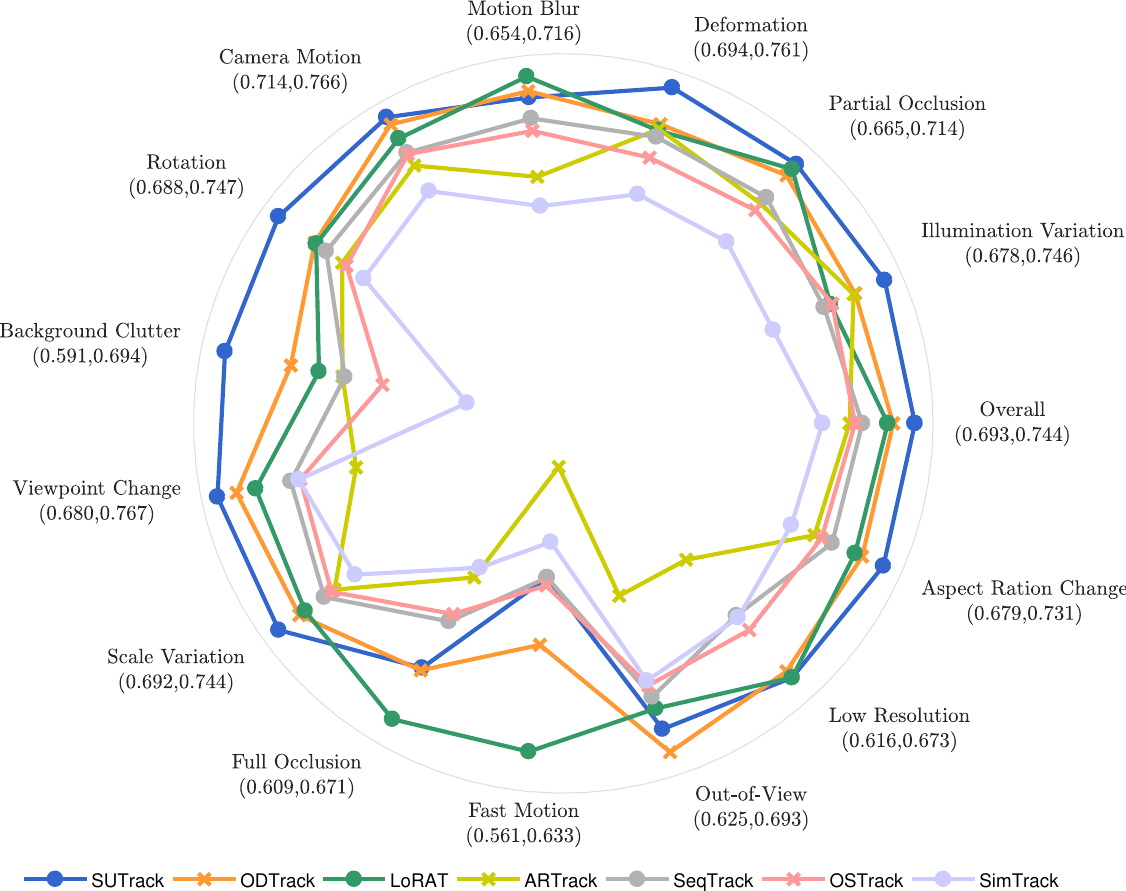}
    \caption{AUC scores of different attributes on LaSOT.}
    \label{fig:lasotattr}
\end{figure}

In Fig.\ref{fig:lasotattr}, we present the attribute-based evaluation results on LaSOT\cite{LaSOT}. Our SUTrack model achieves the best performance in most attributes, particularly excelling in background clutter, deformation, illumination variation, and rotation, which require robust appearance modeling. This underscores SUTrack's strong capabilities as a foundational unified tracking model. However, we observe that SUTrack performs less effectively in motion-related attributes, such as fast motion. 
This is due to two factors: first, we use a relatively small search region, and second, we have not integrated advanced motion and temporal information modeling techniques~\cite{odtrack,artrackv2}. We plan to investigate this in future work.

\section{Additional Ablation Study}
In this section, we provide additional ablation studies. We use SUTrack as the baseline model, and its results are reported in Tab.~\ref{tab-ablation-supp} (\#1).

\subsection{Auxiliary Modality}
We conduct experiments to investigate the impact of auxiliary modalities on performance in four multi-modal SOT tasks: RGB-Depth, RGB-Thermal, RGB-Event, and RGB-Language tracking. 
Specifically, we remove the auxiliary modality input and use only the RGB modality for tracking.
The results shown in \#2 of Tab.~\ref{tab-ablation-supp} indicate a decline in performance, confirming that our default model effectively utilizes auxiliary modalities to enhance tracking performance.

\subsection{Token for Task-recognition}
As discussed in the main manuscript, we compute the average of all feature embeddings output by the transformer model to produce a single vector for the task-recognition prediction. 
We also explore two alternatives: using the text token embedding or an additional task token embedding for this prediction. The results, shown in \#4 and \#5 of Tab.~\ref{tab-ablation-supp}, indicate that these methods perform slightly inferior to our default averaging approach. A potential reason is that the default averaging method offers more direct supervision for all feature embeddings. Moreover, all these methods perform better than the approach in Tab.~\ref{tab-ablation-supp} (\#3) that does not use the task-recognition training strategy, validating the effectiveness of this strategy.

\subsection{Initialization of the multi-modal patch embedding parameters}

As described in the ``More Implementation Details" section of this appendix, we divide the pre-trained patch embedding parameters by a factor of two, and load them into the first three and last three channels of the multi-modal patch embedding in the SUTrack model.
This ensures that the output value range remains consistent with the pre-trained model. We refer to this approach as the half-copy method.
Additionally, we explore two alternatives:
i) Full-copy method: We load the pre-trained patch embedding parameters into the first three and last three channels of the multi-modal patch embedding without dividing them by two. The results are reported in Tab.~\ref{tab-ablation-supp} (\#6). 
ii) Single-copy method: We load the pre-trained patch embedding parameters into only the first three channels of the multi-modal patch embedding, while the last three channels are randomly initialized. 
The experimental results are shown in Tab.~\ref{tab-ablation-supp} (\#7). The results show that our default half-copy method delivers the best performance, confirming its effectiveness.

\section{Limitation}

One limitation of SUTrack is that, although it addresses a wide range of existing SOT tasks, its generalization to potential new SOT tasks is unknown. In our ablation studies, we demonstrate that SUTrack possesses some zero-shot generalization ability on the current SOT tasks, but its capacity to handle new tasks that may arise in the future remains uncertain. A potential solution to this limitation could be the development of continual learning methods to enhance SUTrack's lifelong learning capabilities.

Additionally, recent tracking algorithms~\cite{odtrack,artrackv2} have demonstrated that incorporating more temporal and motion information can significantly enhance the performance of base trackers. However, since this work focuses on providing a foundational unified tracking model, we have not yet explored techniques for modeling temporal and motion information. Consequently, the model's full potential may not have been fully realized.


\bibliography{aaai25}

\end{document}